# Adaptive control of resource flow to optimize construction work and cash flow via online deep reinforcement learning

Can Jiang [a, b], Xin Li [a], Jia-Rui Lin [b*], Ming Liu [a], Zhiliang Ma [b]

*[a] Glodon Company Limited, Beijing, China, 100193*
*[b] Department of Civil Engineering, Tsinghua University, Beijing, China, 100084*
*\* Corresponding author: lin611@tsinghua.edu.cn, jiarui_lin@foxmail.com*

**Abstract**

Existing approaches in construction project management failed to address the integrated control of resource flow. Therefore, this study proposes a novel model and method to adaptive control the resource flows to optimize the work and cash flows of construction projects. First, a mathematical model based on a partially observable Markov decision process is established to formulate the complex interactions of construction work, resource, and cash flows as well as uncertainty and variability of diverse influence factors. Meanwhile, to efficiently find the optimal solutions, a deep reinforcement learning (DRL) based method is introduced to realize the continuous adaptive optimal control of labor and material flows, thereby optimizing the work and cash flows. To assist the training process of DRL, a simulator based on discrete event simulation is also developed to mimic the dynamic features and external environments of a project. Experiments in simulated scenarios illustrate that our method outperforms the vanilla empirical method and genetic algorithm, possesses remarkable capability in diverse projects and external environments, and a hybrid agent of DRL and empirical method leads to the best result. The proposed model and method may serve as a step stone for adopting DRL technology in construction project management.

## 1. Introduction

Since the most important project metrics relate to the work and cash flows, to optimize these two flows is usually the objective of construction project management. The work flow represents the progressions of diverse trades (a trade represents the work content for a specified type of workers) [1], whose metrics include delays on trades and durations. The cash flow comprises cash inflow (e.g., receiving milestone payment from clients) and outflow (e.g., payment for staff wages and materials ordering) [2], whose metrics include net present value (NPV) and costs.

Labor and material flows are two of the most essential resource flows, and critical to construction project. Poor management on labor and material flows causes unstable work flows and negative cash flows, further leads to delays [3] and wastes [4] in construction project, eventually time or/and cost overrun [5]. It is necessary to support project managers with optimization approaches for resource-flow-control.

However, existing optimization approaches in project management failed to address the integrated control of resource flow. First, existing optimization models limited the scope of their decision variables to avoid modelling the extremely complex and uncertain environments. Most existing models are based on transformation or flow view [6]. The decision variables of most transformation-based models are the start times of activities, whereas the influence of resource flows is represented by constraints; most flow-based models [7] failed to comprehensively consider all resource flows, i.e., they considered only labor or material flow.

Second, even if the desired model is proposed, it is difficult to solve this extremely complex and uncertain model with the conventional methods. The high dimensionality and large intervals of decision variables reflect its complexity; this is difficult to handle using the evolutionary algorithm (EA) [8] or traditional reinforcement learning (RL) [9] method. In contrast, existing approaches such as rule-based methods cannot address the uncertainty, and therefore, it is difficult for them to perform well under multiple stochastic variables.

Therefore, this study proposed an adaptive and continuous resource-flow control model to optimize the work and cash flows of construction projects. The decision variables in the model include the numbers of allocated work hours and quantities of ordered materials. Our model outperforms the existing models in considering the complexity and uncertainty. Regarding the complexity, it considers the various factors which affect the work or cash flow. For example, factors such as the quantity of the work content, precedence of activities, productivity of workers, and material availability affect the work flow [10], whereas staff wages, quantity of material ordering, payment method, etc. influence the cash flow of the project [11]. Regarding the uncertainty, our model considers the variability in the influencing factors of the work or cash flow [12]. For example, the variability of weather condition and workforce motivation [13] causes the uncertainty of work flow, and the variability of the payment schedule and unit price of resources causes the uncertainty of cash flow [14].

Moreover, we propose the proximal policy optimization (PPO) [15], an online deep reinforcement learning (DRL) method [16], to solve our model. The extraordinary capabilities of DRL algorithms in handling continuous adaptive optimal control problems in complex and dynamic environments has been proven [17]; DRL-based agents make decisions based on their observations of state gradually, whereas conventional methods like EA output the whole decision trajectory based on the initial status of the project. DRL methods have an additional advantage, that is the flexibility of DRL algorithms allow us to propose hybrid agents, and such agents may lead to better result.

The remainder of this paper is organized as follows. In Section 2, we review the existing mathematical models and compare them with our model. Section 3 introduces necessary knowledge on the features of the DRL algorithms. Section 4 presents the model formulation and Section 5 describes the methodology to find the optimal solution. In Section 6, we evaluate the performance of our methodology with numerical experiments; finally, this study is concluded in Section 7. More details on our model are provided in Appendices A, B, and C.

## 2. Literature review on existing mathematical models

Transformation and flow views are the two main views in time-cost optimization models on project management [6]. The transformation-based models represent construction projects as dependencies between activities; they consider the optimization of the duration and cost of the project as schedule problems, i.e., the decision variables are start times of all activities. Flow-based models not only divide a construction project into activities; however, they assume that the operation of an activity requires specific flows (i.e., labor and material flows) to be ready [7]. These models optimize the total duration and cost by controlling and leveling the flows. Fig. 1 illustrates the differences between these two views with an example.

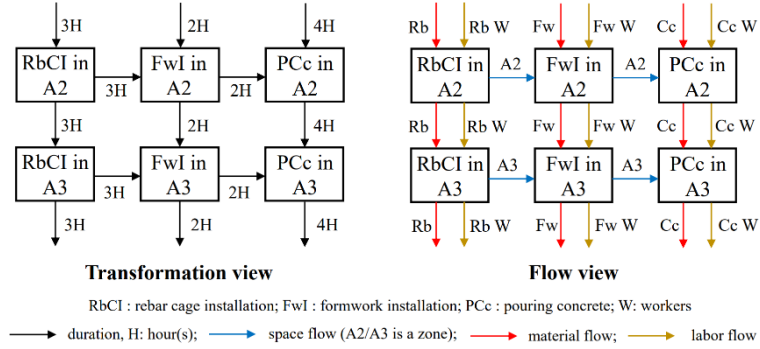

Fig. 1. Transformation versus flow views

The rest of this section reviews the transformation and flow-based optimization models in Sections 2.1 and 2.2, respectively. Models integrating both of these two views are introduced in Section 2.3, and Section 2.4 compares the existing models with our model.

**2.1. Transformation-based models**

The typical transformation-based optimization models are resource-constrained project-scheduling problems (RCPSPs) [18]. The basic RCPSP model minimize the duration and/or cost of the project by arranging activities with precedence relations, and they assume that each activity need a fixed duration and resources usage. These resources are usually renewable resources, i.e., crews and machines; they have fixed capabilities during the project.

Researchers have proposed enhanced versions based on the basic model. Wang et al. [19] proposed a multimode RCPSP; a mode indicates a specific operation of an activity and each mode corresponds to a specific cost and duration of an activity. Liu and Wang [20] and He et al. [21] enhanced the multimode RCPSP. The former study considered cash flows as a constraint whereas the basic model assumed an infinite startup budget; the latter study planed the payment schedule under the constraint of cash flow. However, both of [20] and [21] ignored the objective of minimizing the duration. Tirkolaee et al. [22] integrated the advantages of these three studies.

Chakrabortty et al. [23] and Sallam et al. [24] considered the uncertainty of the activity duration by assuming that the durations obey the probability distributions; the latter study guaranteed the robustness of the outputted schedule by limiting the probability of activity delays. Prayogo et al. [25] considered the minimization of the quantities of necessary renewable resources as the objective of scheduling, whereas Ma et al. [26] considered the quantities of available crews and machines as decision variables. Wang et al. [27] considered the transfer costs of resource allocation; Alcaraz et al. [28] assumed the resource costs are time-dependent.

**2.2. Flow-based models**

Flow-based models optimize time and cost by controlling the labor, material, and space flows. The labor-flow-control models schedule work and the idle time of workers; its basic model is look-ahead planning (LAP) [29]. The material-flow-control models plan the ordering quantities or the baseline of inventory stock, and its basic models include the economic-order quantity [30] and construction logistics planning (CLP) [31]. The basic models of space-flow-control include location-based management system [32] and Takt time planning (TTP) [33].

Automatic optimization methodologies and enhanced versions of these models are proposed in recent years. For labor-flow-control models, Al-Rawi and Mukherjee [34] and Soman and Molina-Solana [35] proposed automatic methods for generating the outputs of LAP. Rahmanniyay and Yu [36] considered multiskilled workers, whereas the basic LAP assumed that each staff must work for a specific trade. For material-flow-control models, Son et al. [37] proposed a simulation-based automatic optimization method for CLP problems. Jaśkowski et al. [38] considered the variability of the unit price of materials and multiple deliver channels, and Kulkarni and Halder [39] treated inventorying leveling and avoiding material shortage as objectives. They considered the uncertainty of the activity duration, and Lu et al. [40] proposed a simulation-based optimization model to determine the level of stock baseline under specific allocation policies. For space-flow-control, Jabbari et al. [41] proposed a model to level the workload of work zones for TTP; the workload leveling results in a minimum Takt time, which reduces the duration of the project.

**2.3. Integrated models**

Some studies considered integrated decisions of project scheduling and flow controlling. Hazır and Schmidt [42] used a schedule model for determining the modes and end times of activities for time optimization; they used a control model to plan work-hour-distribution of each activity to save money. Hosseinian et al. [43] proposed a model for handling both schedule and human resource allocation problems; they assumed the inexperienced workers can learn from the skilled ones while working together. Almatroushi et al. [44] proposed a hybrid model for scheduling the noncritical activities and order-consumable resources; however, the schedule of critical activities remains fixed.

**2.4. Summary**

The proposed models do not comprehensively consider the complexity and uncertainty of construction projects. The managers need to create a large amount of decisions for optimizing the work and cash flows to account for complexity. However, the proposed models consider only the limited number of decision variables; many important influencing factors need to be inputted, or disregarded. Table 1 indicates that the transformation-based models only determine activity schedules without planning the usages of workforce and materials; the flow-based and integrated models focus only on single-flow control. Almost half of the existing models have a single objective. The other half of the models aim to achieve time-cost trade-off optimization; they optimize costs without considering the possibility of broken cash flow.

Table 1. Decision variables, constraints, and objectives of each optimization model

| Study | | Decision variable or constraint | | | | | | Objective | |
|---|---|---|---|---|---|---|---|---|---|
| | | Activity schedule | Payment schedule | Crew number | Work time | Material ordering | Space usage | Time | Cost |
| Transformation-based model | [19] | ★ | | ▲ | | | | ○ | ○ |
| | [20] | ★ | ★ | ▲ | | | | | ○ |
| | [21] | ★ | ★ | ▲ | | | | | ○ |
| | [22] | ★ | ★ | ▲ | | | | ○ | ○ |
| | [23] | ★ | | ▲ | | | | ○ | |
| | [24] | ★ | | ▲ | | | | ○ | |
| | [25] | ★ | | ★ | | | | | |
| | [26] | ★ | | ★ | | | | ○ | ○ |
| | [27] | ★ | | ▲ | | | | ○ | ○ |
| | [28] | ★ | | ▲ | | | | ○ | |
| Integrated model | [42] | ★ | | ▲ | ★ | | | ○ | ○ |
| | [43] | ★ | | ▲ | ★ | | | ○ | ○ |
| | [44] | ▲/★ | | ▲ | | ★ | | | ○ |
| Flow-based model | [34] | ★ | | ▲ | ★ | | | | ○ |
| | [35] | ★ | | ▲ | ★ | | | ○ | ○ |
| | [36] | ▲ | ▲ | ▲ | ★ | | | ○ | ○ |
| | [37] | ▲ | ▲ | | | ★ | | | ○ |
| | [38] | ▲ | ▲ | | | ★ | | | ○ |
| | [39] | ★ | | | ★ | ★ | | ○ | ○ |
| | [40] | ★ | | | | ★ | | ○ | ○ |
| | [41] | ★ | | ▲ | | | ★ | ○ | |
| | Ours | ★ | ★ | ▲ | ★ | ★ | ▲ | ○ | ○ |

* ★ represent decision variables or depend on decision variables, ▲ represent constraints

There are diverse sources of uncertainty in actual construction projects. However, as indicated in Table 2, nearly half of the existing models do not consider uncertainty at all, and many studies only assume that activity durations are random numbers. Few studies consider uncertainty from other sources such as human resource and material; the modelling of the internal mechanism of uncertainty remains rare. Therefore, we manage to propose a novel optimization model that can outperform the existing models when considering the complexity and uncertainty.

Table 2. Uncertainty in each optimization model

| Study | | Time | | Human resource | | Material | | External Factor | |
|---|---|---|---|---|---|---|---|---|---|
| | | Activity duration | Lead times | Unit cost | Efficiency | Unit Cost | Waste | Weather | Market |
| Transformation-based model | [19] | ▲ | | ▲ | ▲ | | | | |
| | [20] | ▲ | | ▲ | | | | | |
| | [21] | ▲ | | ▲ | | | | | |
| | [22] | ▲ | | ▲ | | | | | |
| | [23] | ★ | | | | | | | |
| | [24] | ★ | | | | | | | |
| | [25] | ▲ | | | | | | | |
| | [26] | ▲ | | ▲ | | | | | |
| | [27] | ★ | ▲ | | | | | | |
| | [28] | ▲ | | ★ | | | | | |
| Integrated model | [42] | ▲ | ▲ | ★ | ▲ | | | | |
| | [43] | ★ | | ★ | ★ | | | | |
| | [44] | ▲ | | ▲ | ▲ | ▲ | ▲ | | |
| Flow-based model | [34] | ★ | | ▲ | ▲ | | | | |
| | [35] | ★ | | ▲ | ▲ | | | | |
| | [36] | ★ | | ▲ | ▲ | | | | |
| | [37] | ▲ | ▲ | | | ★ | | | ▲ |
| | [38] | ▲ | ▲ | | | ★ | ▲ | | ★ |
| | [39] | ★ | ★ | | | ▲ | ★ | | |
| | [40] | ★ | ▲ | | | ▲ | | | |
| | [41] | ▲ | | | ▲ | | | | |
| | Ours | ★ | ▲/★ | ★ | ★ | ★ | ★ | ★ | ★ |

* ★ represent random numbers or depend on random numbers, ▲ represent deterministic

## 3. Background on deep reinforcement learning

Project management with flow-based models is an optimal control problem proposed to design a resource-flow control agent for optimizing work and cash flows. Recent studies in DRL technology have proven its capability for solving optimal control problems. In the term DRL, "deep" represents agents that are on the basis of deep neural networks (DNN); "reinforcement learning" (RL) represents agents that learn optimal strategies by trial and error in specific environments [16]. The control problems must be formulated as a MDP to adopt DRL algorithms [16], e.g., a partially observable Markov decision process (POMDP) [45], which is not difficult for flow-based models.

The rest of this section is organized as follows. Section 3.1 introduces the POMDP, and Section 3.2 compares EA, RL, and DRL, and it illustrates the advantages of online DRL. Section 3.3 shows the flexibility of the DRL algorithms. In Section 3.4, we describe the training and validation of the agents.

**3.1. Partially observable Markov decision process**

Fig. 2 shows that the POMDP models the process wherein an agent changes the state of an environment by taking actions. The agent takes action based on its observation.

$$action_t = \text{Policy}(observation_t), \quad (1)$$

where Policy() represents the logic of the agent's decision making. The action changes the state of the environment.

$$state_{t+1} = \text{Transition}(state_t, action_t), \quad (2)$$

where Transition() represents the mechanism of the environment. A new observation is obtained based on the historical record of observable state parameters.

$$observation_{t+1} = \text{Observe}(SRecord_{t+1}), \quad (3)$$

where $SRecord_{t+1} = \{state_1, state_2, \ldots, state_{t+1}\}$. The reward represents the quantitative evaluation of the state change, and

$$r_t = \text{Reward}(state_{t+1}, state_t, action_t), \quad (4)$$

where $r_t$ is known to the agent in model-based RL, while it is unknown in the model-free RL.

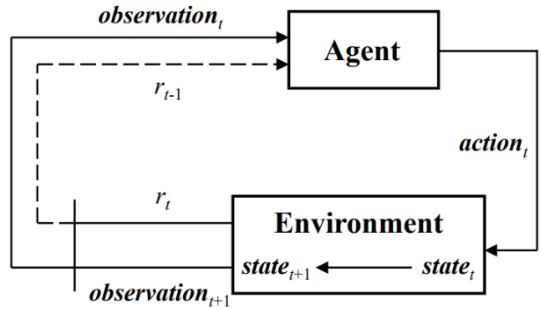

Fig. 2. Basic framework for partially observable Markov decision process

If the initial state is given,

$$state_t = \mathbf{init\_state}. \quad (5)$$

Therefore, the trajectories of observation, action, state, and reward can be calculated by repeatedly calling Eq. (3) → Eq. (1) → Eq. (2) → Eq. (4).

The objective of POMDP is to maximize the cumulative reward for the entire process by finding the optimal policy function of the agent.

$$\text{Policy}^*() := \max \sum_{t=1}^{T} r_t; \quad (6)$$
$$s.t. \ (2), (4), (5)$$

**3.2. Comparison among EA, RL, and DRL**

The EA and RL methods are adopted to solve the optimization models reviewed in Section 2. For example, a symbiotic organisms search [46], a EA method, was adopted in [26] to solve the optimal activity schedule and crew numbers. In [24], the authors used a RL-EA hybrid methods to solve RCPSPs;

two EA methods optimized the activity schedule and Q-learning [9], an RL algorithm, was determined in which the EA method was used in a specific problem. In [35], the authors determined the workforce allocation on each day, and Q-learning was used. In [40], the authors adopted genetic algorithm (GA) [8], which is a EA method, to solve optimal material ordering.

However, our POMDP-based model has considerably more decision variables compared to the optimization models in these studies. EA, RL, and even offline DRL cannot handle the complexity. The decision variables of a POMDP model represent the action trajectory, {*action*$_1$, ..., *action*$_t$, ..., *action*$_T$}. The EA methods directly solve the optimal trajectory, whereas RL and DRL divide the problem into similar sub-problems; they output each *action*$_t$ based on *observation*$_t$. Assume that the numbers of possible time steps, observations, and actions are T, M and N, respectively. The EA methods can handle this optimization problem when all of them are small; the RL algorithms, when T is large but M and N are small; the offline DRL algorithms, when T, M is large but N is small; and the online DRL algorithms, when all of them are large.

EA methods like GA convert the action trajectory to a binary chromosome code whose length is T · $\log_2$N; the convergence becomes difficult when the chromosome code is extremely long. RL algorithms like Q-learning need to build a Q-table, whose element at the $m^{th}$ row and $n^{th}$ column represents the expected total cumulative reward for taking the $n^{th}$ action under the $m^{th}$ observation; it is almost impossible to update the Q-table if M or N is very large.

The DRL comprises offline and online DRL. Deep Q-learning [47] is an offline DRL method whose architecture of the policy function is illustrated in Fig. 3(a). A normalization module first converts *observation*$_t$ to $o^t$; the DNN(s) map $o^t$ to $q^t$, whose $n^{th}$ element ($q_n^t$) represents the expected total cumulative reward for taking the $n^{th}$ action under current observation. A SoftMax function converts $q^t$ to {$p_1^t$, $p_2^t$, ......, $p_N^t$}, a discrete probability distribution, which is used to sample $a^t$. Finally, a denormalization module converts $a^t$ to *action*$_t$. The size of the output layer of the DNN(s) is N; the convergence is difficult if N is very large.

The PPO is an online DRL algorithm, whose architecture of the policy function is shown in Fig. 3(b). The outputs of the DNN(s) are $v^t$ and *mean*$_a^t$ instead of the Q-values; $v^t$ reflects the evaluation of the agent of total cumulative reward under *observation*$_t$; *mean*$_a^t$ is used to sample $a^t$ according to the Gaussian distribution $N($*mean*$_a^t$, $\exp($*logstd*$_a))$. The recommended action is determined after sampling and denormalization.

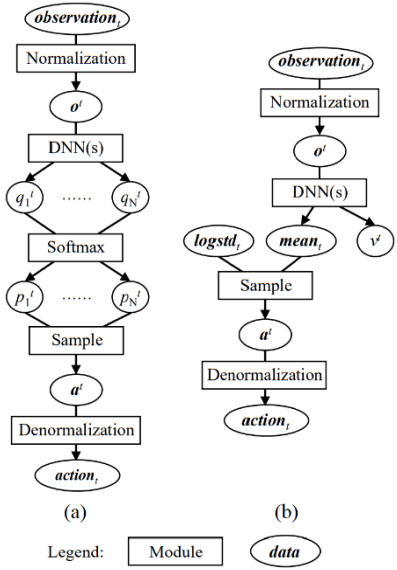

Fig. 3. Architecture of the policy function when (a) offline and (b) online DRL algorithms are used

### 3.3. Flexibility of DRL algorithms

A DRL-based agent can adopt multiple DNNs to perform part of actions or cooperate with rule-based methods; such strategies can lead to better results. Oroojlooyjadid et al. [48] simulated the beer supply chain and trained four DNNs with deep Q-learning, each of which corresponds to an individual decision maker in the supply chain. Any trained DNN can cooperate with others or ruler-based policies and perform well. Chen et al. [49] investigated the optimal energy management policy of a building cluster, and they evaluated different approaches using CityLearn [50] and observed that an approach combining the DRL and rule-based policies performed the best.

### 3.4. Training and validating DRL

Unlike deep learning, which collects training data from the real world, the DRL usually train agents with data generated by simulators. The DRL algorithms were first adopted in abstract strategy games such as Go [51]; then, they were used to play computer and video games such as StarCraft 2 [52]. All of them are problems in the virtual world. Although the DRL was used to solve real-world problems such as robot navigation [53], the training in the real world is rare because of the cost and potential risk of trial and error. Many simulation platforms [54] have been proposed to support agent training; however, there are more or less bias between the simulated and real environments, which is called the reality gap. Some researchers attempted to propose approaches to narrow the reality gap [55]; more developers attempt to let people believe that the reality gaps of their simulators are acceptable by making the software or even the technical details available for the potential users.

Validating is easier than training in the real world; however, building a testbed is expensive, and the potential risk still existed. The simulated scenarios are considered by researchers to evaluate the capabilities of their DRL agents. In few research studies, the authors designed simulated task scenarios by themselves; then, they selected non-DRL-based approaches as baselines. Hu et al. [56] developed DRL-based agents to control unmanned aerial vehicles; they designed virtual tasks to compare the proposed DRL and dynamic programming approaches. Further, many benchmark libraries were built to support the validation; each library includes simulated scenarios for a specific type of task. For example, AI2-THOR [57] is used for visual-based indoor robot navigation.

## 4. Mathematical model for adaptive resource-flow control

We formulate the adaptive resource-flow control problem in a construction project as a POMDP. Section 4.1 defines this engineering problem, explains what each component of the POMDP represents in the problem, and discusses the complexity and uncertainty of our model. Sections 4.2–4.7 describe the formulation of the components of the POMDP in detail, respectively.

### 4.1. Map the engineering problem to the POMDP

#### 4.1.1. Problem definition

The aim of the adaptive resource-flow control problem is to determine the work-hour allocation and material ordering for each day; the decision variables are thus **WH** and **B**, which are sequences of allocated work hours and brought materials, respectively. The basic goal is to avoid the project's failure caused by the break in cash flow or delay in work flow; the advanced goal is to maximize the NPV and minimize the duration of the project, respectively. The objective can be represented by

$$\text{Maximize: } f(T, Ca, A), \qquad (7)$$

where $T$, $Ca$, and $A$ represent sequences corresponding to time, holding cash, and progress.

The controlling is constrained by the laws of nature and construction project, and this can be briefly represented by Eqs. (8)–(15).

$$g_1(T, A) = 0, \tag{8}$$

which means the duration of projects depends on the progress.

$$g_2(Ca, A, WH, B, Pr) = 0, \tag{9}$$

where $Pr$ represents the sequence corresponding to the material price. The cash inflow depends on the progress, whereas the cash outflow depends on work hours allocated and materials bought.

$$g_3(E, WH, We) = 0, \tag{10}$$

where $E$ and $We$ represent the sequences corresponding to productivity of the worker and the weather. Fatigue and bad weather conditions can reduce the productivity.

$$g_4(A, WH, E, S) = 0, \tag{11}$$

where $S$ represents the sequence corresponding to the material stock. The update of the progress is based on the critical flow theory [58]; the labor, material, and precedence flows can be considered the critical flow.

$$g_5(S, A, B) = 0. \tag{12}$$

Thus, the consumption of material depends on the progress, and the replenishment of material depends on the bought material.

$$Ca \geq g_6(WH, B, Pr), \tag{13}$$

which implies holding enough money to pay labor and material costs.

$$4 \leq WH \leq 12, \tag{14}$$

which implies the work hours of the crews are more than 4 and less than 12 on each day.

$$0 \leq B \leq B_{max}, \tag{15}$$

which implies the quantity of material purchased has an upper limit. The upper limit usually is the maximum storage capacity.

**4.1.2. Assumptions and limitations of our model**

Diverse sorts of construction project obey the constraints described in Section 4.1.1, but these constraints are expressed in different mathematical form for different sort of project. In order to clearly define the mathematical expression of our model, we consider the following four preliminary assumptions:

A.1. The project's goal is constructing a multistory building, and each floor of this building has the same area and construction space zoning; the area of the zones is the same.

A.2. The zones of each floor are numbered, and they must be constructed in the order of their numbers.

A.3. There are the following three types of onsite activities: rebar cage installation, formwork installation, and pouring concrete; three types of materials and workers, i.e., rebar/formwork/concrete (workers), correspond to these types of activities, respectively.

A.4. The precedence of construction activities is rebar cage installation → formwork installation → pouring concrete; formwork installation or pouring concrete can only be performed in the zones where the corresponding prior activity has been completed, and rebar cage installation can only be performed in zones whose downstairs zones were completed (concrete has been completed poured).

These assumptions determine the forms of constraints and parameters. For example, A.3 implies that the decision parameters, $B$ and $WH$, correspond to three types of material and workers, and A.1, A.2 and A.4 detail the influence of the precedence flow in Eq. (11). However, they also limit the scope of our model, which only apply to a specified sort of project, i.e., multistory reinforced concrete buildings. Further, our model does not address the influence of construction machine/equipment.

**4.1.3. Representations of the components of POMDP**

Formulating the problem as a POMDP is a prerequisite for solving it with a DRL algorithm. Basic components of a POMDP include action, state, transition, observation, policy, and reward, as indicated in Section 3.1.

Although decision variables in Eq. (6) are internal parameters of the policy function, the actions are outputs of the policy function. The actions represent decisions to control the labor and material flows, i.e., ***WH*** and ***B***, which are detailed in Section 4.2. The states represent the status of flows in the project and external environment, e.g., ***Ca***, ***A***, and ***Pr***, which are detailed in Section 4.3. The transition function mimics the laws of nature and the construction project; this is described in Section 4.4. This function comprises the cash, labor, work and material flow modules that correspond to Eqs. (9)–(12), respectively. The observation function mimics human managers collecting necessary information from historical record of the states, which is introduced in Section 4.5. The policy function mimics human managers taking decisions based on the observations; this is introduced in Section 4.6. The rewards reflect the objectives, i.e., Eq. (7), which is detailed in Section 4.7.

**4.1.4. Complexity and uncertainty of the problem**

Our model outperforms existing models in considering the complexity and uncertainty of construction projects. The complexity is reflected by actions and transition functions. For the actions, our model can handle three types of workers and materials simultaneously; all these actions are in large intervals. For comparison, study [35] determined only the workforce allocation, and their decision variables are binaries, each of which represent whether each type of workers work on each day. In [40], the authors proposed a model whose decision variables correspond to the ordering of only one type of materials. The transition function considers complex interactions among flows shown in Fig. 4 and Eqs. (9)–(12). For example, Eq. (9) reflects the cash flow is influenced by work, labor, material, and price flows.

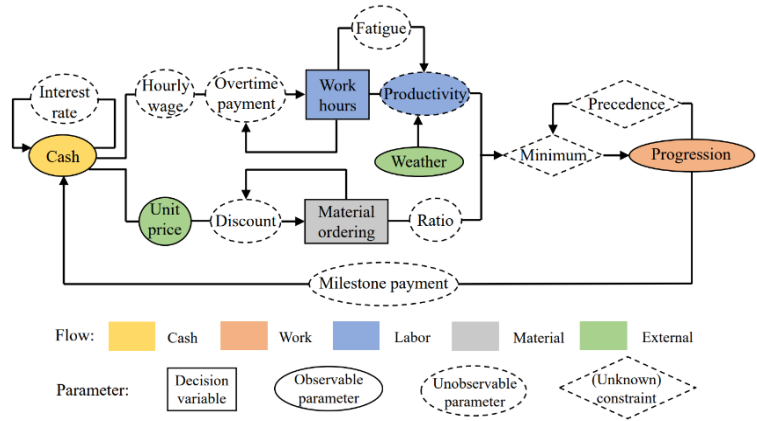

Fig. 4. Interactions among influencing factors

The uncertainty is reflected by the transition function, and it considers the uncertainty from time and cost incurred by modelling the deviations of their underlying factors. The activity duration is affected by the productivity of workers, and it is affected by the weather; the material cost is affected by the unit price that depends on the market. The variability of the weather and the market are modelled as random processes.

**4.2. Actions**

In our model, the trajectory {***action***$_1$, …, ***action***$_t$, …, ***action***$_T$} represents the decisions to control the labor and material flows. Here, ***action***$_t$ represents the work-hour allocation and material ordering based on the $t^{th}$ day, whose parameters are defined in Table 3.

Table 3. Definitions of decision variables ($action_t$)

| | Parameter | Definition | Flow | Interval | Unit |
|---|---|---|---|---|---|
| $WH_t$ | $RbWH_t$ | Work hours for rebar workers | Labor | [4,12] | hour |
| | $FwWH_t$ | Work hours for formwork workers | | | |
| | $CcWH_t$ | Work hours for concrete workers | | | |
| $B_t$ | $RbB_t$ | Quantity of rebars bought | Material | [0, 500] | 0.1 tons |
| | $FwB_t$ | Quantity of formwork bought | | [0, 2000] | square meter |
| | $CcB_t$ | Quantity of concrete bought | | [0, 300] | cubic meter |

### 4.3. State

The state in our model reflects the status of the work, cash, labor, material, and external flows. The parameters of $state_t$ are summarized in Table 4. The parameters that correspond to the labor flow are unobservable; the others are observable state parameters. As described in Eq. (5), the $state_1$ should be given, and this represents the status at the beginning of the project. The initial values of all state parameters are listed in Table 4, wherein InitCa and SDate are the input parameters. InitCa represents the amount of start-up cash for the project and SDate represents the start date of the project in this calendar year; modules in the transition function can calculate the initial price and weather states according to SDate.

Table 4. Definitions of state parameters ($state_t$)

| | Parameter | Definition | Flow | Initial value |
|---|---|---|---|---|
| $A_t$ | $RbA_t$ | Total area where rebar cages are installed | Work | 0 |
| | $FwA_t$ | Total area where formworks are installed | | 0 |
| | $CcA_t$ | Total area where concrete is poured | | 0 |
| | $Ca_t$ | Amount of holding cash | Cash | InitCa |
| | $ICa_t$ | Amount of incoming cash | | 0 |
| | $LWSa_t$ | Cumulative amount of last week's salary | | 0 |
| | $EnE_t$ | Productivity reduction coefficient caused by bad weather | Labor | 1 |
| $FaI_t$ | $FaRbI_t$ | Long-term fatigue index for rebar workers | | 0 |
| | $FaFwI_t$ | Long-term fatigue index for formwork workers | | 0 |
| | $FaCcI_t$ | Long-term fatigue index for concrete workers | | 0 |
| $S_t$ | $RbS_t$ | Quantity of rebars stock | Material | 0 |
| | $FwS_t$ | Quantity of formworks stock | | 0 |
| | $CcS_t$ | Quantity of concrete stock | | 0 |
| | $FwU_t$ | Quantity of formworks in use | | 0 |
| $Pr_t$ | $RbPr_t$ | Unit price of rebar | External | Correspond to SDate |
| | $FwPr_t$ | Unit price of formwork | | |
| | $CcPr_t$ | Unit price of concrete | | |
| | $Tp_t$ | Temperature | | |
| | $Rf_t$ | Rainfall | | |
| | $Ws_t$ | Wind speed | | |

### 4.4. Transition

The transition function, Eq. (2), defines the update mechanism of $state_t$, and this contains modules that correspond to work, cash, labor, material, and external flows, respectively. The remainder of this subsection discusses the modules of the transition function; the corresponding model parameters are defined in Table 5.

Table 5. Definitions of model parameters

| Parameter | | Definition | |
|---|---|---|---|
| TF | | Number of floors in the building | Basic project information |
| pFZ | | Number of construction zones on a floor | |
| pZA | | Area of construction zone | |
| MaxT | | Maximum project duration | |
| pFCa | | Amount of milestone payment for each floor completed | Relate to the cash flow |
| IR | | Daily interest rate to hold cash | |
| pHSa | pRbHSa | Hourly salary for the normal working hours of rebar workers | |
| | pFwHSa | Hourly salary for the normal working hours of formwork workers | |
| | pCcHSa | Hourly salary for the normal working hours of concrete workers | |
| OWSaR | | Extra percentage of hourly salary for overtime work | |
| pIMCa | | Daily fee for material inventory management | |
| DcB | DcRbB | Quantity of rebars to be bought to avail the minimum discount | |
| | DcFwB | Quantity of formworks to be bought to avail the minimum discount | |
| | DcCcB | Quantity of concrete to be bought to avail the minimum discount | |
| MinDcR | | Minimum discount to buy materials | |
| W | RbW | Number of rebar workers | Relate to the labor flow |
| | FwW | Number of formwork workers | |
| | CcW | Number of concrete workers | |
| NAbR | | Normal absence ratio of workers | |
| MaxS | MaxRbS | Maximum storage capacity of rebars | Relate to the material flow |
| | MaxFwS | Maximum storage capacity of formworks | |
| FwRcLR | | Average loss ratio of recycling the formworks | |
| pWH2A | pRbWH2A | Average area that a rebar worker can complete per normal work hour | Relate to the work flow |
| | pFwWH2A | Average area that a formwork worker can complete per normal work hour | |
| | pCcWH2A | Average area that a concrete worker can complete per normal work hour | |
| pS2A | pRbS2A | Average area that a unit of rebar can complete | |
| | pFwS2A | Average area that a unit of formwork can complete | |
| | pCcS2A | Average area that a unit of concrete can complete | |

**4.4.1. Cash-flow module**

This module focuses on the update of the cash flow, and this consists of inflow and outflow.

$$Ca_{t+1} = Ca_t + ICa_t - OCa_t. \qquad (16)$$

The inflow comprises a milestone payment from the client and interest from holding money; milestone payments are payments at event occurrences [59], and the manager applies to the client for a fixed payment after a floor is completed; the lead time is 2–4 days.

$$ICa_t = \begin{cases} \text{IR} \cdot Ca_{t-1} + \text{pFCa}, & t \in \textbf{\textit{PayDays}} \\ \text{IR} \cdot Ca_{t-1}, & t \notin \textbf{\textit{PayDays}} \end{cases}, \qquad (17)$$

where ***PayDays*** represents the dates when the milestone payment was received.

The outflow comprises a payment for the salaries of the workers and the cost of material ordering and storage. The payments for salaries of the workers are progress payments [59]; the workers receive their salaries once a week until the project is completed, and the amount of a payment depends on last week's working hours. The payments for the material ordering are lump-sum payments [59], and the entire payment is paid while ordering.

$$OCa_t = \begin{cases} \text{pIMCa} + BPa_t + LWSa_t, & t = 1 \pmod{7} \\ \text{pIMCa} + BPa_t, & t \neq 1 \pmod{7} \end{cases}, \tag{18}$$

where $BPa_t$ represents today's payment for material ordering. $BPa_t$ depends on the quantities of material ordering, unit price, and discount ratio.

$$BPa_t = \sum DcR_t \cdot Pr_t \cdot B_t. \tag{19}$$

The relationship between the $DcR_t$ and $B_t$ is shown in Eq. (20) based on the investigation of Min and Pheng [60].

$$DcR_t = \begin{cases} 1 - \dfrac{B_t}{\text{DcB}} \cdot (1 - \text{MinDcR})_t, & 0 \leq B_t \leq \text{DcB} \\ \text{MinDcR}, & B_t > \text{DcB} \end{cases}. \tag{20}$$

The salaries for the last week are given by

$$LWSa_t = \sum_{t'=t-7}^{t-1} \sum DSa_{t'}, \tag{21}$$

where $DSa_t$ represents the wage for rebar, formwork, or concrete workers on the $t^{\text{th}}$ day. The $DSa_t$ depends on the number of attendance workers ($AW_t$), work hours, and hourly salary for the normal work hours and overwork hours. Working hours over 8 hours are considered overtime; the hourly wage for overtime working is 1+OWSaR times the normal hourly wage.

$$DSa_t = \begin{cases} AW_t \cdot \text{pHSa} \cdot WH_t, & WH_t \leq 8 \\ AW_t \cdot \text{pHSa} \cdot [WH_t + \text{OWSaR} \cdot (WH_t - 8)], & WH_t > 8 \end{cases}. \tag{22}$$

$AW_t$ depends on the absence ratio ($AbR_t$) as shown by

$$AW_t = W_t \cdot (1 - AbR_t). \tag{23}$$

where $AbR_t$ is positively correlated with the fatigue index $FaI_t$, which represents that fatigue causes a high absence ratio [61].

### 4.4.2. Labor-flow module

This module focuses on the update of the productivity of the workers. The productivity reduction is caused by physical and environmental reasons based on the study reported in [61]; fatigue caused by working overtime and poor weather conditions are critical physical and environmental reasons, respectively.

The influence of fatigue is represented by $FaI_t$ and $FaE_t$; the fatigue index, $FaI_t$, depends on work hours, and the overtime works ($WH_{t-1} > 8$) can increase $FaI_t$ as shown in

$$FaI_t = \max(0, 0.5 FaI_{t-1} + WH_{t-1} - 8). \tag{24}$$

$FaE_t$ represents the productivity reduction coefficient caused by long-term fatigue; it is a piecewise linear and monotonic decreasing function of its corresponding $FaI_t$, and $FaE_t = 1$ when $FaI_t = 0$.

The influence of weather is represented by $EnE_t$.

$$EnE_t = \min(1, EnE_{t-1} + 0.3, EnTpE_t, EnRfE_t, EnWsE_t), \tag{25}$$

where $EnTpE_t$, $EnRfE_t$, and $EnWsE_t$ represent the effects of temperature, rainfall, and wind speed, respectively. $EnRfE_t$ and $EnWsE_t$ are piecewise linear and monotonic decreasing functions of $Rf_t$ and $Ws_t$, respectively. $EnTpE_t$ is a piecewise linear function of $Tp_t$, and $EnTpE_t$ is very small when $Tp$ is extremely high or low.

The detailed expressions of these four piecewise linear functions are set by users of our model; Table 6 shows an example of the expressions of these functions; they are recommended by human project managers.

Table 6 Example of the detailed expressions of piecewise linear functions

| Breakpoints of the piecewise linear function $FaE_t$ |
|---|

| Value of $FaE_t$ | 1 | 0.95 | 0.85 | 0.7 | 0.4 |
|---|---|---|---|---|---|
| Value of $FaI_t$ | 0 | 2 | 4 | 6 | 8 |
| Breakpoints of the piecewise linear function $EnTpE_t$ ||||||
| Value of $EnTpE_t$ | 0 | 0.5 | 1 | 1 | 0.5 | 0 |
| Value of $Tp_t$ (°C) | 0 | 10 | 20 | 30 | 40 | 50 |
| Breakpoints of the piecewise linear function $EnRfE_t$ ||||||
| Value of $EnRfE_t$ | 1 | 0.8 | 0.5 | 0 |
| Value of $Rf_t$ (mm) | 2 | 10 | 20 | 50 |
| Breakpoints of the piecewise linear function $EnWsE_t$ ||||||
| Value of $EnWsE_t$ | 1 | 0.7 | 0 |
| Value of $Ws_t$ (m/s) | 5 | 10 | 20 |

### 4.4.3. Work-flow module

The work-flow module focuses on the update of the project's progression. The work content, i.e., the effort required for completing a construction project, is proportional to the area of the corresponding construction space; we thus measure the progression of the project by the area of the completed construction space ($A_t$).

The update of progressions is based on the critical flow theory [58]. The increments of progressions ($\Delta A_t$), i.e., completed work contents, on each day are influenced by the combination of labor, material, and precedence flows; the bottlenecks for the entire process are called critical flows. For example, if the manager makes the rebar workers work for a long time but only buys a few rebars, the material flow is the critical flow; if the manager buys several materials and makes the workers work for a long time but the previous activity is not performed, the critical flow is precedence flow. $\Delta A_t$ is calculated by

$$\Delta A_t = \min(MaxWA_t, MaxMA_t, MaxPA_t, \text{TA} - A_t), \tag{26}$$

where $MaxWA_t$, $MaxMA_t$, or $MaxPA_t$ represents the completed work contents when the labor, material, or precedence flow is the critical flow, respectively. The item $\text{TA} - A_t$ indicates that the completed areas cannot exceed the total construction area of the building (TA).

$$\text{TA} = \text{TF} \cdot \text{pFZ} \cdot \text{pZA}. \tag{27}$$

$MaxWA_t$ is positively related to the total effective working hours, and the productivity reduction coefficients ($FaE_t$ and $EnE_t$)

$$MaxWA_t = (1+\delta) \cdot FaE_t \cdot EnE_t \cdot \text{pWH2A} \cdot AW_t \cdot EWH_t, \tag{28}$$

where $EWH_t$ represents the effective work hours of the $t^{\text{th}}$ day. $EWH_t$ equals to the cumulative effective time of all work hours.

$$EWH_t = \sum_{i=1}^{WH_t} \text{pEWH}_i. \tag{29}$$

Alvanchi et al. [62] observed that short-term fatigue and darkness in the evening reduce productivity. Thus, we define $\text{pEWH}_i = 1$ when $i \leq 8$; $\text{pEWH}_i$ is a monotonically decreasing sequence of $i$ when $i \geq 8$. Further, $\delta$ represents a random variable that obeys $U[-0.05, 0.05]$, which indicates that a worker's productivity is random instead of a deterministic parameter.

$MaxMA_t$ depends on the material stock and ratio between the constructed area and consumed material; the randomness of the ratio is also considered.

$$MaxMA_t = (1+\delta) \cdot \text{pS2A} \cdot S_t. \tag{30}$$

$MaxPA_t$ is related to the current progression. The activities, formwork installation, and concrete pouring can be performed in construction zones where the corresponding last process was completed yesterday.

$$MaxFwPA_t = \left\lfloor \frac{RbA_t}{\text{pZA}} \right\rfloor \cdot \text{pZA} - FwA_t, \tag{31}$$

$$MaxCcPA_t = \left\lfloor \frac{FwA_t}{pZA} \right\rfloor \cdot pZA - CcA_t. \tag{32}$$

The activity rebar-cage installation can be performed only in zones whose downstairs zones were completed yesterday.

$$MaxRbPA_t = (\left\lfloor \frac{CcA_t}{pZA} \right\rfloor + pFZ) \cdot pZA - RbA_t. \tag{33}$$

**4.4.4. Material-flow module**

The material-flow module focuses on updating the stock of the materials. The stock of the material ($S_t$) changes because of the outflow and inflow. The material inflow comprises ordered and recycled materials; the lead time of the material deliver is 1 day, and the material outflow comprises consumed and wasted materials. The waste is attributed to the storage capacity.

The rebar is storable but not recyclable, and the formwork is storable and recyclable; their stock cannot exceed the maximum storage capacity (MaxS). The concrete is not storable; the concrete not consumed today will be discarded, and the concrete available tomorrow is the concrete ordered today.

$$S_{t+1} = \begin{cases} \min(MaxS, S_t - C_t + B_t), & \text{rebar} \\ \min(MaxS, S_t - C_t + Rc_t + B_t), & \text{formwork} \\ B_t, & \text{concrete} \end{cases} \tag{34}$$

where $C_t$, $Rc_t$, and $B_t$ represent the quantities of the material consumed, recycled, and bought on $t^{th}$ day. Further, $C_t$ is positively related to the completed area ($\Delta A_t$).

$$C_t = (1+\delta) \cdot \frac{\Delta A_t}{MaxMA_t} \cdot S_t. \tag{35}$$

The formwork will be removed if the concrete is poured. The quantity of the removed formwork ($FwM_t$) is proportional to the area of the zones completed today.

$$FwM_t = \frac{(\left\lfloor \frac{CcA_{t+1}}{pZA} \right\rfloor - \left\lfloor \frac{CcA_t}{pZA} \right\rfloor) \cdot pZA}{FwA_t - \left\lfloor \frac{CcA_t}{pZA} \right\rfloor \cdot pZA} \cdot FwU_t. \tag{36}$$

The quantity of the formwork in terms of the used and recycled concrete can be calculated by following Eqs. (37) and (38), respectively.

$$FwU_{t+1} = FwU_t + FwC_t - FwM_t; \tag{37}$$

$$FwRc_t = [1 - (1+\delta) \times FwRcLR] \times FwM_t, \tag{38}$$

where FwRcLR represents the loss ratio, which indicates that not all the removed formworks can be recycled. Further, $\delta$ represents a random variable that obeys $U[-0.1, 0.1]$, which indicates that the loss ratio is a random parameter.

**4.4.5. External-flows module**

The external-flows module focuses on the updates of weather and material prices, and they do not rely on action. First, we build random processes based on historical statistics in the real world; then, we sample the values of weather and price states based on built processes. The technical details of this module are described in Appendix A.

**4.5. Observations**

As described in Eq. (3), the observation function mimics collecting necessary information for decision making in real projects; the necessary information is ***observation***$_t$. In our model, the ***observation***$_t$ comprises (1) date information, i.e., $t$; (2) observable state parameters, i.e., parameters in Table 5 except $EnE_t$ and $FaI_t$; (3) historical weather information within the last 2 days, i.e., $Tp_{t-2}/Rf_{t-2}/Ws_{t-2}$ to $Tp_{t-1}/Rf_{t-}$

$_{t-1}/Ws_{t-1}$; (4) historical work hours within the last 3 days, i.e., $WH_{t-3}$ to $WH_{t-1}$; (5) forecasts of cash inflow within the next 3 days, i.e., predicted values from $ICa_{t+1}$ to $ICa_{t+3}$; and (6) forecasts of weather information and material prices.

Observe() comprises the selection and forecast modules. The selection module picks up the first four components of **observation**$_t$ from **SRecord**$_t$, and the forecast module mimics the forecast of cash inflow, weather, and material prices in the real world. The technical details of the forecast module are described in Appendix B.

### 4.6. Policy

The policy function represents the resource-flow control agent in our model, which output **action**$_t$ according to **observation**$_t$. Our model is more complex than existing ones, and **action**$_t$ and **observation**$_t$ have too many elements as shown in Section 4.2 and Section 4.5. Assume that the number of possible actions and observations are N and M; if we treat any element of **action**$_t$ or **observation**$_t$ as a continuous variable; N or M are infinite. Even if we treat all these elements as discrete variables, N and M are large because of the high dimensionality and large intervals of **action**$_t$ and **observation**$_t$.

As described in Section 3.2, the policy function can be based on a traditional RL method when both of M and N are small; based on an offline DRL method when M is huge but N is small; and based on an online DRL method when both M and N are large. Thus, our policy function is based on an online DRL method, i.e., the PPO algorithm; its architecture is shown in Fig. 3(b).

### 4.7. Reward

The reward function is adopted to evaluate the degree of the resource-flow control agent that achieves our objective. The objective is the optimization of cash and work flows; this comprises the following four sub-goals: (1) avoiding project failure caused by cost overrun, i.e., $Ca_t \geq OCa_t$ (the remaining money is sufficient for salary payment and material ordering); (2) avoiding the project failure caused by a time overrun, i.e., $T \leq \mathrm{MaxT}$; (3) maximize the profit ($NPV_T$); and (4) minimize the duration of the entire projects ($T$).

However, rewards associated with these sub-goals are sparse, and they can only be obtained in very few successful or failure states. The agent can rarely reach the success states at the beginning of the training. If we use only sparse rewards, the agent cannot identify which attempts are closer to the success states; further, they cannot make progress. According to Ng et al. [63], dense rewards are required to guide the agent towards the success states. Thus, the reward function must consider the dense and sparse rewards that correspond to each sub-goals, as shown in

$$r_t = \omega_\mathrm{P} \cdot DeR_\mathrm{P}^t + \omega_{\mathrm{D},1} \cdot DeR_\mathrm{D}^t + \omega_{\mathrm{W},1} \cdot DeR_\mathrm{W}^t + \omega_{\mathrm{M},1} \cdot DeR_\mathrm{M}^t \\ + \omega_\mathrm{F} \cdot SpR_\mathrm{F}^t + \omega_{\mathrm{D},2} \cdot SpR_\mathrm{D}^t + \omega_{\mathrm{W},2} \cdot SpR_\mathrm{W}^t + \omega_{\mathrm{M},2} \cdot SpR_\mathrm{M}^t , \qquad (39)$$

where $\omega$s represent the weights. The users of our model can determine the precedence of the sub-goals by themselves, and they set the values of $\omega$s according to the precedence.

Further, $DeR_\mathrm{P}^t$ and $SpR_\mathrm{F}^t$ represent the dense and sparse rewards related to sub-goals (1) and (2), i.e., avoiding project failure. $DeR_\mathrm{P}^t$ guides the agent in advancing its progression, which equals the ratio of the area completed on the $t^\mathrm{th}$ day.

$$DeR_\mathrm{P}^t = \frac{(CcA_t - CcA_{t-1})}{\mathrm{TA}} , \qquad (40)$$

where $SpR_\mathrm{F}^t$ represents the negative reward for project failure.

$$SpR_\mathrm{F}^t = \begin{cases} -1, & Ca_t < OCa_t \text{ or } t > \mathrm{MaxT} \\ 0, & \text{otherwise} \end{cases} . \qquad (41)$$

Further, $DeR_\mathrm{D}^t$ and $SpR_\mathrm{D}^t$ correspond to sub-goals (2) and (4), i.e., the optimization of the work flow.

The agent receives negative rewards when the duration of the project is close to the maximum value that the client can tolerate.

$$DeR_D^t = \begin{cases} 0, t < 0.7\text{MaxT} \\ -\dfrac{1}{\text{MaxT}}, 0.7\text{MaxT} \leq t < 0.85\text{MaxT} \\ -\dfrac{4}{\text{MaxT}}, 0.85\text{MaxT} \leq t < \text{MaxT} \end{cases}. \quad (42)$$

$$SpR_D^t = \begin{cases} \min(0, \dfrac{t-1}{\text{MaxT}} - 0.7), CcA_t = \text{TA} \\ 0, CcA_t < \text{TA} \end{cases}. \quad (43)$$

$DeR_W^t$, $DeR_M^t$, $SpR_W^t$, and $SpR_M^t$ correspond to sub-goals (1) and (3), i.e., the optimization of the cash flow. $DeR_W^t$ and $SpR_W^t$, and $DeR_M^t$ and $SpR_M^t$ represent dense and sparse rewards corresponding to labor and material expenses, respectively. The $NPV_T$ is the most critical index of the cash flow, and this equals the cumulative cash inflow minus outflow.

$$NPV_T = Ca_T - Ca_1 = \sum_{t=1}^{T}(ICa_t - OCa_t). \quad (44)$$

The cumulative cash outflow consists of the cumulative labor and material expenses, i.e., $WCost_T$ and $MCost_T$.

$$\sum_{t=1}^{T} OCa_t = WCost_T + MCost_T. \quad (45)$$

The main part of the cash inflow is the milestone payment.

$$\sum_{t=1}^{T} ICa_t \approx \text{TMilePay} = \text{TF} \cdot \text{pFCa}. \quad (46)$$

Thus, $DeR_W^t$ and $DeR_M^t$ are adopted to minimize the daily labor and material expenses, respectively.

$$DeR_W^t = Wratio \cdot (\dfrac{Ca_t - Ca_{t-1}}{\text{TMilePay}}), \quad (47)$$

$$DeR_M^t = Mratio \cdot (\dfrac{Ca_t - Ca_{t-1}}{\text{TMilePay}}), \quad (48)$$

where $Wratio$ and $Mratio$ represent the approximate ratios of $WCost_T$ and $MCost_T$ in the project's total cost.

If the agent profits by cutting labor and material costs, it obtains $SpR_W^t$ and $SpR_M^t$ when the project is completed.

$$SpR_W^t = \begin{cases} \max(0, Wratio \cdot \text{TMilePay} - WCost_t), CcA_t = \text{TA} \\ 0, CcA_t < \text{TA} \end{cases}. \quad (49)$$

$$SpR_M^t = \begin{cases} \max(0, Mratio \cdot \text{TMilePay} - MCost_t), CcA_t = \text{TA} \\ 0, CcA_t < \text{TA} \end{cases}. \quad (50)$$

The agent will receive a huge negative reward when it is just learning to complete the building (it has not learned to save money) if Eqs. (49) or (50) do not contain the maximum functions. In this case, the agent prefers to maintain 99% progression rather than complete the entire project.

## 5. Model solving based on the online DRL

Fig. 5 describes the procedures to solve the optimal resource-flow control problem. Five resource-flow control agents are proposed, and each of them is based on one or two DNNs. The architectures of these DNNs are determined in Section 5.1. In Section 5.2, we present the developed discrete event simulator (DES) based on the resource-flow control model. The training data are generated by calling the DES in Section 5.3, and we use the PPO algorithm to train the agents to learn the optimal control policies in Section 5.4.

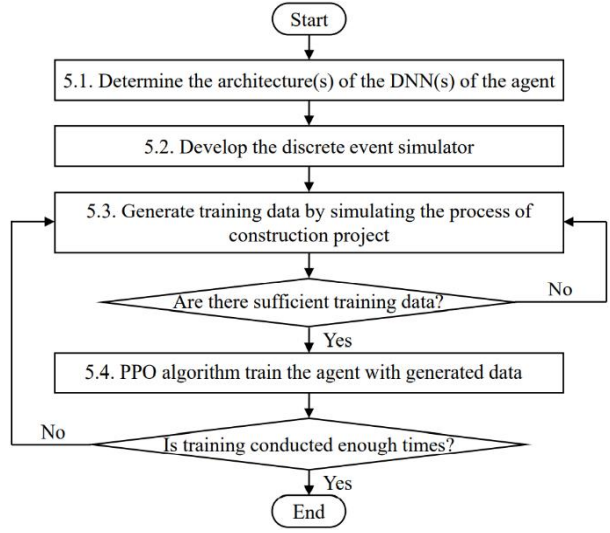

Fig. 5. Procedures to solve the resource-flow control problem

**5.1. Determine the architecture(s) of the DNN(s)**

Fig. 3(b) shows the architecture of the policy function of an agent. The normalization module first converts ***observation**_t* to $o^t$; one or two DNNs map $o^t$ to $v^t$ and ***mean**_a^t*, $v^t$ reflects the evaluation of the agent's current status of the project (the criteria is the reward function), and ***mean**_a^t* is used to sample $a^t$ based on the Gaussian distribution $N($***mean**_a^t$, \exp($***logstd**_a$))$; $a^t$ corresponds to the recommended decisions; i.e., ***action**_t*, and a denormalization module that converts $a^t$ to ***action**_t*.

The DNN adopted in our policy function is a multilayer fully connected network, which contains a basement ($B_\xi$), value ($V_\varphi$), and policy ($\pi_\theta$) network. The $B_\xi$s and $V_\varphi$s of the five proposed agents share the same architecture, whereas there are three types of architectures for their $\pi_\theta$s: full, work hour, and material. These architectures are presented in Fig. 6, where the number in the lower right corner of each rectangle represent the size of the corresponding output vector. All activation functions in these networks are rectified linear units, except the last hidden layers of the policy networks, whose activation functions are hyperbolic tangents (tanh). According to Lillicrap et al. [64], batch normalization rarely improves the capability of a DRL agent; the hidden layers of our networks do not contain batch normalization units.

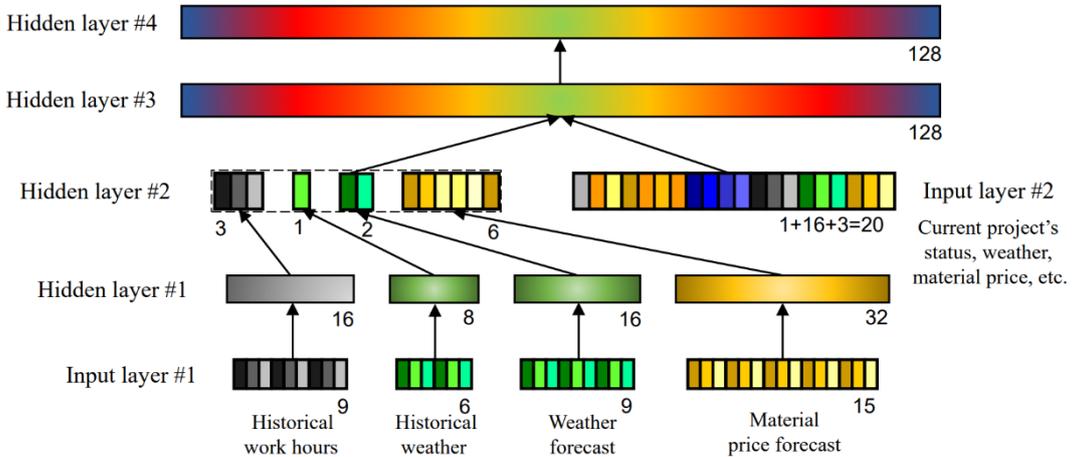

(a)

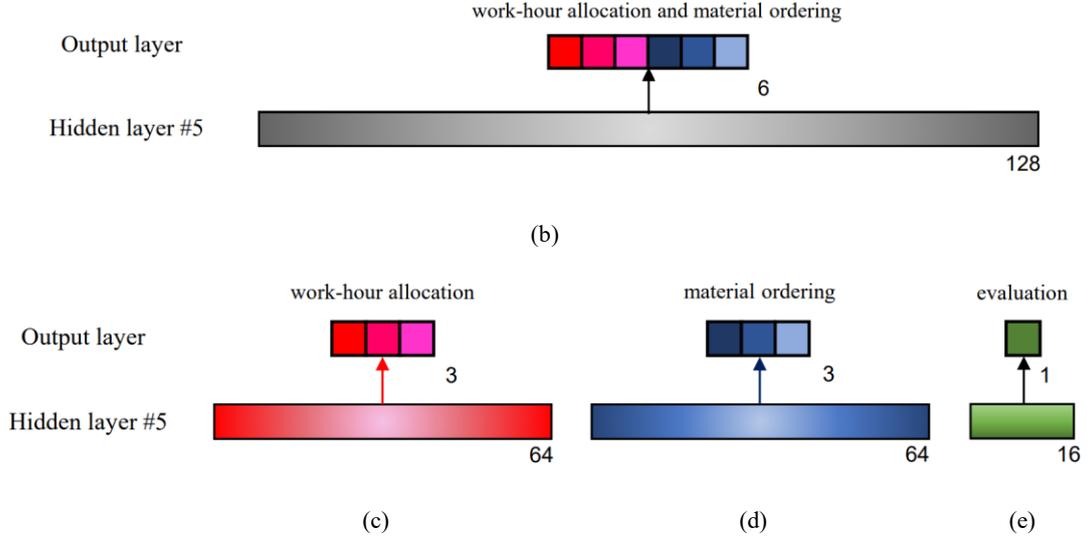

Fig. 6. Architecture of the (a) basement network, (b) full, (c) work hour, (d) material policy network, and (e) value network.

The input of $B_\xi$ is $o^t$; the parameters are distributed in different input layers based on their influence on decision making. The parameters that are indirectly influenced are first processed by the 1$^{st}$ and 2$^{nd}$ hidden layers, and then, a feature vector with a size of 12 is output. Subsequently, this vector is concatenated with parameters that are directly influenced and processed by the 3$^{rd}$ and 4$^{th}$ hidden layers. The output of $B_\xi$ is thus produced, and its size is 128; this vector is the input to $V_\varphi$ and $\pi_\theta$; $v^t$ and **mean**$_a^t$ are the outputs, respectively. Further, $v^t$ indicates that the expected cumulative reward can be obtained if the agent takes the best actions every time afterward. The output **mean**$_a^t$ of the full policy network (Fig. 6(b)) corresponds to decision variables for labor and material flows management. The work hour (Fig. 6(c)) or material (Fig. 6(d)) policy networks only makes decisions for labor or material flow management, respectively.

**5.2. Develop the DES**

We develop a DES based on our proposed resource-flow-control model. Our proposed model is a POMDP model that consists of the policy, transition, observation, and reward functions shown in Section 3.1. The DES includes modules corresponding to these functions.

The policy function of a resource-flow control agent can be pure or hybrid; the architecture of pure DRL-based policy functions is illustrated in Fig. 3(b). Section 3.3 introduces the advantages of the hybrid policies of DRL and rule-based methods. Therefore, we propose five agents with pure or hybrid policy functions as shown in Fig. 7. The policy functions of the 1$^{st}$ and 2$^{nd}$ agents share the same architecture as shown in Fig. 7(a), i.e., they adopt a single full policy network (SFPN) to manage the labor and material flows; their reward functions are different. The 3$^{rd}$ agent uses single work-hour policy network (SWPN) to manage the labor flow; the material flow is controlled by a rule-based policy as shown in Fig. 7(b). The situation for the 4$^{th}$ agent is the opposite of 3$^{rd}$ one as shown in Fig. 7(c), i.e., single material policy network (SMPN) for material flow and rule-based policy for labor flows. The rule-based policies in the 3$^{rd}$ and 4$^{th}$ agents are halves of an empirical policy; this mimics the mind of human managers. The technical details of the empirical policy are described in Appendix C. The 5$^{th}$ agent adopts a double-policy network (DPN) architecture, and this contains two DNNs, as shown in Fig. 7(d). All these agents are implemented in Python3 with the PyTorch library.

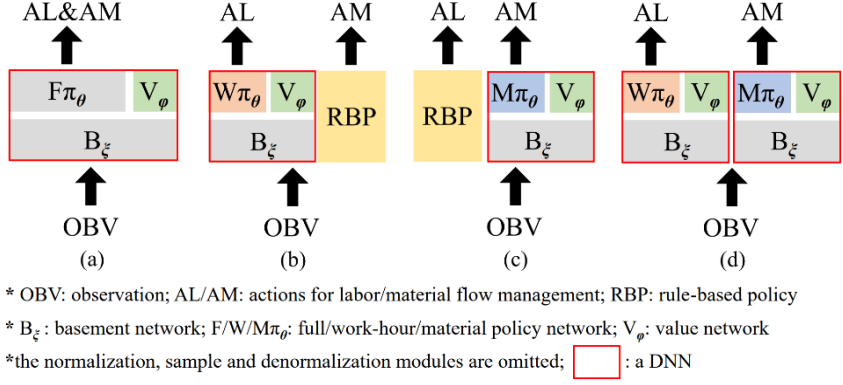

Fig. 7. Policy functions for five agents: (a) SFPN for the 1st and 2nd agents; (b) SWPN for the 3rd agent; (c) SMPN for the 4th agent; and (d) DPN for the 5th agent

The formulation of the reward function is described in Section 4.7. As indicated in Eq. (39), the reward function comprises some components; their weights need to be set with the following three principles: (1) in terms of the precedence of the sub-goals: if saving the labor cost is more important than saving the material cost, $\omega_{W,1}$ and $\omega_{W,2}$ are higher than $\omega_{M,1}$ and $\omega_{M,2}$; (2) balancing the weights of dense and sparse rewards: if the weights for sparse rewards are higher, the training will be more difficult; however, the performance of the agent will be better; (3) the value of the cumulative reward should be in the interval [0, 1] that benefits the convergence of the DNNs. A DNN corresponds to a reward function, and the weights for the first four agents (which contain only one DNN) are summarized in Table 7. The 5th agent is a DPN architecture that contains the DNNs of the 3rd and 4th agents and their corresponding reward functions.

Table 7 Weights of the reward functions of the agents

|  | $\omega_P$ | $\omega_{W,1}$ | $\omega_{M,1}$ | $\omega_{D,1}$ | $\omega_{W,2}$ | $\omega_{M,2}$ | $\omega_{D,2}$ | $\omega_F$ |
|---|---|---|---|---|---|---|---|---|
| Agent #1: SFPN | 0.5 | 0.25 | 0.25 | 0.5 | 2 | 2 | 1 | 0.15 |
| Agent #2: SFPN | 0.5 | 1 | 0.125 | 0.5 | 8 | 1 | 1 | 0.15 |
| Agent #3: SWPN | 0.5 | 2 | 0 | 0.5 | 16 | 0 | 1 | 0.15 |
| Agent #4: SMPN | 0.5 | 0 | 0.25 | 0.5 | 0 | 2 | 1 | 0.15 |

The formulations of the transition and observation are described in Section 4.4 and 4.5, respectively; all of them are implemented in Python3. The transition module is adopted to simulate the construction project and external environment; we validated its correctness by unit tests with real project data; the results show that the reality gap is acceptable.

### 5.3. Generate the training data

The DRL methods solve the optimal resource-flow-control actions by optimizing the policy function. As described in Section 3.2 and 5.1, the internal parameters of the policy function include $\theta$, $\varphi$, $\xi$, and $logstd_a$; the training data are used to update these parameters.

The process of a construction project is simulated by repeatedly calling the DES, and the training data are collected simultaneously, as illustrated by the pseudocode in Table 8. The state and its historical record are initialized in lines 1 and 2, and they corresponds to Eq. (5). Line 3 determines whether the simulated project is the end; $CcA_t < TA$ implies that the project is yet to be successfully completed; the second and third conditions, $t < MaxT$ and $Ca_t \geq ICa_{t-1}$, indicate that the project has failed because of the time and cost overruns, respectively. Mention that,

$$Ca_t \geq ICa_{t-1} \Leftrightarrow Ca_{t-1} \geq OCa_{t-1}. \tag{51}$$

Line 5 calls the observation function, i.e., Eq. (3); lines 6–9 call the policy function, i.e., Eq. (1); line 10 calls the transition function, i.e., Eq. (2); and line 11 calls the reward function, i.e., Eq. (4). In line 12, P($a^t$ | $o^t$, $\pi_\theta$) represents the possibility of the DNN recommending $a^t$ when the input is $o^t$; this can be calculated using Eq. (52).

$$P(a^t \mid o^t, \pi_\theta) = \frac{1}{\sqrt{2\pi} \cdot \exp(logstd_a)} \exp\{-\frac{[a^t - \pi_\theta(B_\xi(o^t))]^2}{2 \cdot [\exp(logstd_a)]^2}\} . \quad (52)$$

A sample of training data is the combination of $o^t$, $a^t$, $v^t$, $r_t$, and P($a^t$ | $o^t$, $\pi_\theta$) collected in line 13. The collected training data are necessary for optimizing the internal parameters of the policy function, i.e., $\theta$, $\varphi$, $\xi$, and $logstd_a$.

Table 8 Pseudocode for generating the training data

| 1 | $t \leftarrow 1$, **SRecord** $\leftarrow$ [] |
|---|---|
| 2 | **state**$_t \leftarrow$ **init_state** |
| 3 | while $CcA_t <$ TA and $t <$ MaxT and $Ca_t \geq ICa_{t-1}$: |
| 4 | **SRecord**.append(**state**$_t$) |
| 5 | **observation**$_t \leftarrow$ Observe(**SRecord**) |
| 6 | $o^t \leftarrow$ Normlization(**observation**$_t$) |
| 7 | **mean**$_a{}^t \leftarrow \pi_\theta(B_\xi(o^t))$, $v^t \leftarrow V_\varphi(B_\xi(o^t))$ |
| 8 | $a^t \leftarrow$ Sample($N$(**mean**$_a{}^t$, $\exp(logstd_a)$)) |
| 9 | **action**$_t \leftarrow$ Denormlization($a^t$) |
| 10 | **state**$_{t+1} \leftarrow$ Transition(**state**$_t$, **action**$_t$) |
| 11 | $r_t \leftarrow$ Reward(**state**$_t$, **state**$_{t-1}$, **action**$_t$) |
| 12 | P($a^t$ \| $o^t$, $\pi_\theta$) $\leftarrow$ Conditional_Prob($a^t$, $o^t$, **mean**$_a{}^t$, $logstd_a$) |
| 13 | **TrainingData**.append([$o^t$, $a^t$, $v^t$, $r_t$, P($a^t$ \| $o^t$, $\pi_\theta$)]) |
| 14 | $t \leftarrow t + 1$ |

## 5.4. Training the agent

Table 9 illustrates the pseudocode of training the agents. As indicated in line 1, the optimization starts after collecting the $T_H$ samples of training data, where $T_H$ is called "the size of the horizon" in the field of DRL. $T_H$ is considerably larger than $T$, i.e., the duration of a simulated project, which indicates that we need to conduct several simulations for collecting $T_H$ samples. **target** and **advantage** are necessary intermediated parameters for optimizing the policy function whose elements include $tar_t$ and $adv_t$. Further, $tar_t$ represents the target value that is a correction of $v^{t+1}$ after considering the feedback of $r_t$; $adv_t$ represents the estimated advantage of $a^t$. Further, $tar_t$ and $adv_t$ are calculated based on $v^t$ and $r_t$ with the generalized advantage estimator [65]. Thus, we calculate $tar_t$ and $adv_t$ by following Eqs. (53)–(56),

$$tar_t = v^t + \sum_{t'=t}^{T_H} I(t,t') \cdot (\gamma \cdot \lambda)^{t'-t} \cdot \delta_{t'} , \quad (53)$$

$$I(t,t') = \begin{cases} 1, & t \text{ and } t' \text{ in the same simulation} \\ 0, & t \text{ and } t' \text{ in the different simulations} \end{cases}, \quad (54)$$

$$\delta_{t'} = r_{t'} + \gamma \cdot v^{t'+1} - v^{t'} , \quad (55)$$

$$adv_t = tar_t - v^t , \quad (56)$$

where both $\gamma$ and $\lambda$ are in the interval [0, 1]. Here, $\gamma$ represents the discount rate for future rewards, and $\lambda$ reflects the bias level of the generalized advantage estimator; a higher value of $\lambda$ lowers the bias but results in a higher variance.

In line 3, we optimize $\theta$, $\varphi$, $\xi$, and $logstd_a$ using the Adam optimizer [66] for $nE$ epochs. Further, we select a batch of samples from the collected $T_H$ samples in each epoch; the batch size is $T_B$. The optimization objective is to minimize a loss function L($\theta$, $\varphi$, $\xi$, $logstd_a$), which reflects the gap between the current DNN and the optimal one. The optimal DNN outputs the action with maximum $adv_t$, and evaluates $v^t$ without deviation. L($\theta$, $\varphi$, $\xi$, $logstd_a$) consists of 3 parts; this is expressed as

$$L(\theta, \varphi, \xi, logstd_a) = L_\pi + c_1 \cdot L_V - c_2 \cdot L_{DistEn}, \tag{57}$$

where $c_1$ and $c_2$ represent the coefficients. The optimization of $L_\pi$ increases the possibility of the policy network ($\pi_\theta$) to select an action that has a higher $adv_t$; however, it needs to avoid the output of $\pi_\theta$ changing too fast.

$$L_\pi = \sum_{t=1}^{T_B}[\min(adv_t \cdot PoR_t, adv_t \cdot \text{clip}(PoR_t, 1-\varepsilon, 1+\varepsilon))], \tag{58}$$

$$PoR_t = \frac{P(a^t|o^t, \pi_\theta)}{P(a^t|o^t, \pi_{\theta,old})}. \tag{59}$$

The optimization of $L_V$ enables the value network ($V_\varphi$) to provide a more precise estimation.

$$L_V = \sum_{t=1}^{T_B}[(tar_t - V_\varphi(B_\xi(o^t)))^2]. \tag{60}$$

$L_{DistEn}$ reflects an entropy bonus for the distribution, $N(mean_a^t, \exp(logstd_a))$.

$$L_{DistEn} = \sum_{t=1}^{T_B}[P(a^t|o^t, \pi_\theta) \cdot \ln(P(a^t|o^t, \pi_\theta))]. \tag{61}$$

The training data is cleared (line 4) after parameter optimization.

Table 9 Pseudocode for training the agent

| 1 | if len(*TrainingData*) = $T_H$: |
|---|---|
| 2 |    *target*, *advantage* ← Advantage_Estimator(*TrainingData*) |
| 3 |    $\theta$, $\varphi$, $\xi$, $logstd_a$ ← Optimizer(*target*, *advantage*, *TrainingData*) |
| 4 |    *TrainingData* ← [] |

## 6. Experiments and results

Decision-making agents learn to manage the labor and material flows of construction projects with our proposed model and methodology. As discussed in Section 3.4, the capabilities of DRL-based agents are validated in simulated scenarios, and non-DRL-based policies need to be the baselines. Experiments are thus conducted, and the rest of this section is organized as follows. Section 6.1 defines the simulated scenarios and training setups adopted in these experiments; Section 6.2 shows the advantages of online-DRL over other optimization methods; Section 6.3 compares the DRL-based agents with the empirical policy used by human managers; Section 6.4 illustrates the performances of DRL-based agents in multiple project setups; Section 6.5 discusses the optimal architecture of the decision-making agent.

**6.1. Simulated scenarios and training setups**

**6.1.1. Scenario #0: real project in Beijing**

Scenario #0 mimics a typical residential project in Beijing; its environmental setups are summarized in Table 10. The constructed building has 25 floors, and the area of each floor is 600 m², which is divided into 12 construction zones. The start date of the project was May 31$^{st}$ (the 151$^{st}$ day of the calendar year) because the following months have good weather for construction. The standard productivity of workers and the ratio between completed area and consumed materials are calculated based on the real records of resource usage and progression. The standard unit prices for different resources are obtained from the financial statement. Based on these prices, we estimate that the cost for each floor does not exceed 350,000 CNY if the empirical optimal policy is adopted. Therefore, we set the amount of a milestone payment (pFCa) equal to 400,000 CNY; the start-up cash is 1 million CNY.

Table 10 Setups for scenario #0

| Parameter | SDate | InitCa | TF | pFZ | pZA | MaxT | pFCa | IR |
|---|---|---|---|---|---|---|---|---|
| Value | 151 | 1,000,000 | 25 | 12 | 50 | 150 | 400,000 | 0.0001 |
| Parameter | RbW | FwW | CcW | NAbR | pRbHSa | pFwHSa | pCcHSa | OWSaR |
| Value | 12 | 20 | 8 | 0.05 | 27.5 | 27.5 | 22.5 | 200% |
| Parameter | pIMCa | MaxRbS | MaxFwS | FwRcLR | DcRbB | DcFwB | DcCcB | MinDcR |
| Value | 1,000 | 500 | 2,000 | 0.05 | 400 | 800 | 150 | 0.9 |
| Parameter | pRbWH2A | pFwWH2A | pCcWH2A | pRbS2A | pFwS2A | pCcS2A | | |
| Value | 2.5 | 1.51 | 3.84 | 1.54 | 0.25 | 3 | | |

**6.1.2. Scenario group #1: change external environmental conditions**

As shown in Table 11, we adjusted the budget, weather, and market conditions for three additional simulated scenarios in scenario group #1, respectively. In scenario #1, we reduced the start-up cash from 1 million CNY to 900,000 CNY. The holding cash reached a minimum in the early stage of the project because of the delay in obtaining the milestone payment; cutting the start-up cash caused the early stage to be more difficult. In scenario #2, the start date of the project was February 1$^{st}$ (the 32$^{nd}$ day of the calendar year). The productivity reduced by 45% on average (whereas the corresponding reduction ratio is 10% in scenario #0) because the extremely low temperatures in February and March prevented workers from working, and the fluctuations in productivity are considerably larger. We increased the number of workers by 65%, and therefore, the budget needed to be increased. In scenario #3, the discount mechanism was eliminated. The theoretical minimal discount for material purchases was 90% in scenario #0.

Table 11 Setups for scenario group #1: different environmental conditions (omit unchanged parameters)

| #0: CC | Parameter | SDate | InitCa | MinDcR | #1: HBC | Parameter | SDate | InitCa | MinDcR |
|---|---|---|---|---|---|---|---|---|---|
| | Value | 151 | 1,000,000 | 0.9 | | Value | 151 | 900,000 | 0.9 |
| #2: HWC | Parameter | SDate | InitCa | MinDcR | #3: HMC | Parameter | SDate | InitCa | MinDcR |
| | Value | 32 | 1,000,000 | 0.9 | | Value | 151 | 1,000,000 | 1 |

* CC, HBC, HWC, and HMC mean common, harsh budget, weather, and market conditions, respectively

**6.1.3. Scenario group #2: change project setups**

As shown in Table 12, we adjusted the quantity of the work content and the number of workers for three additional simulated scenarios in scenario group #2, respectively. In scenario #4, we increased the number of floors (of the constructed building) and the maximum project duration by 20%, respectively. In scenario #5, we increased the area of each floor from 600m$^2$ to 750m$^2$. The number of workers also increased by 25% to ensure the makespan unchanged, and the milestone payment for a completed floor increased by 25% too. In scenario #6, we changed the number of workers for each type in different proportions.

Table 12 Setups for scenario group #2: different projects (omit unchanged parameters)

| #0: RPB | Parameter | TF | pFZ | pZA | MaxT | pFCa | RbW | FwW | CcW |
|---|---|---|---|---|---|---|---|---|---|
| | Value | 25 | 12 | 50 | 150 | 400,000 | 12 | 20 | 8 |
| #4: INF | Parameter | TF | pFZ | pZA | MaxT | pFCa | RbW | FwW | CcW |
| | Value | 30 | 12 | 50 | 180 | 400,000 | 12 | 20 | 8 |
| #5: IAF | Parameter | TF | pFZ | pZA | MaxT | pFCa | RbW | FwW | CcW |
| | Value | 25 | 15 | 50 | 150 | 500,000 | 15 | 25 | 10 |
| #6: CNWDP | Parameter | TF | pFZ | pZA | MaxT | pFCa | RbW | FwW | CcW |
| | Value | 25 | 12 | 50 | 150 | 400,000 | 16 | 16 | 9 |

* RPB, INF, IAF, and CNWDP mean real project in Beijing, increasing the number of floors, increasing the area of

each floor, and changing the number of workers in different proportions, respectively

**6.1.4. Training setups**

The training setups are summarized in Table 13. We determined the hyperparameters and coefficients of the loss function by considering the experience of other DRL studies. We trained DRL agents on a computer with an i7-11700K @3.60GHz CPU and Nvidia RTX 3060 GPU, and saved the values of the agent's ($\theta$, $\varphi$, $\xi$, and $logstd_a$) and normalization parameters (means and standard deviations of $observation_t$) at every 20 updates.

Table 13 Hyper-parameters and coefficients of loss function

| Parameter | $T_H$ | $T_B$ | $nE$ | $\gamma$ | $\lambda$ | Learning rate | $\varepsilon$ | $c_1$ | $c_2$ |
|---|---|---|---|---|---|---|---|---|---|
| Value | 1024 | 128 | 16 | 0.99 | 0.95 | 0.0001 | 0.2 | 0.5 | 0.01 |

We trained three agents and selected the best one, then tested it five times and showed its average performance as the result of each experimental group. We optimized with GA five times then showed its best performance, or tested the empirical policy five times then showed its average performance as the result of each control group.

**6.2. Comparison between online DRL and GA**

**6.2.1. Experimental design**

DRL-agents with an SFPN architecture were trained and tested in scenario #0, and a GA method was the baseline. This experiment included two control groups with different runtimes; the runtime of GA was approximately equal to the training time of online DRL in the first one, and much longer than in the second one.

According to Section 3.2, we can infer that RL and offline DRL cannot be the baselines because of the high dimensionality and large intervals of the decision variables. RL algorithms like Q-learning need to build a M×N Q-table, while offline DRL like deep Q-learning need a DNN whose output layer's size is N. M and N is the numbers of possible observations and actions respectively; they are infinite in our problem because the observation and action spaces are continuous. Even if we discretize them, M and N will not be small because the dimensions are too high. (Dimensions of observation and action spaces is 59 and 6, respectively.)

**6.2.2. Result and discussion**

The GA can optimize and simulate more times in the same runtime than the online DRL, because it need not to update the DNN. However, its performances are far less well than the online DRL as shown in Table 14. In control group #1, we set the number of generations and population size to 1024 and 256, respectively, to ensure the runtime of GA was approximately equal to the training time of online DRL. However, the GA method did not complete the project, and cost overrun occurred on the 50th day. We increased the runtime by approximately seven times in control group #2, but the performance of GA has barely improved. The reason was that the length of the chromosome code was approximately 2000 in this problem, which made the convergence difficult.

Table 14 Comparison between online DRL and GA

| Method | Number of optimization | Number of simulation | Runtime (minute) | Reward | Progression | Performance |
|---|---|---|---|---|---|---|
| GA | 1024 | 262144 | 18.4 | 0.08 | 64.5% | Cost overrun occurs on the 50th day |
| GA | 10000 | 2000000 | 140.4 | 0.19 | 86.3% | Cost overrun occurs on the 69th day |
| Online | 500 | 9000 | 14.7 | 1.09 | 100% | Complete the project in average 89 |

| DRL | | | | | | day and profit |
|---|---|---|---|---|---|---|

Training an agent with the online DRL method required 14.7 minutes. Fig. 8 shows the reward of an DRL-based agent during the training process, the agents outperforms the GA method after approximately 60 optimizations, breaks even after approximately 80 optimizations and outperforms the empirical policy after approximately 360 optimizations. Further, the trained agent can complete the project in average 89 days and profit.

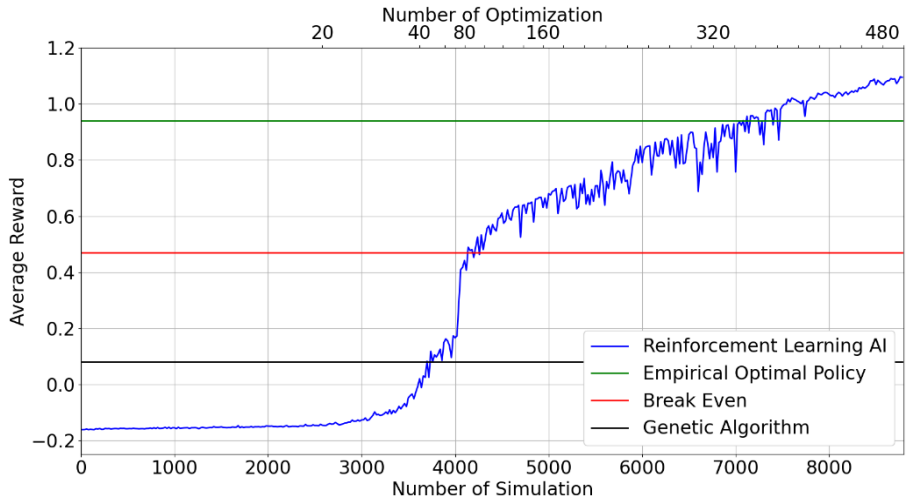

Fig. 8. Reward of a DRL-based agent during training progress

### 6.3. Comparison between the DRL-based and empirical methods

#### 6.3.1. Experimental design

An empirical policy used in the real project was selected as the baseline, which is the static rule-based method discussed in Appendix C. We trained DRL-agents with an SFPN architecture in scenario group #1, which means this experiment included four experimental and control groups, and they illustrated the performances of DRL-agents and the empirical policy under common and harsh budget, weather, and market conditions, respectively.

It is difficult to design a dynamic rule-based policy with high capability. As an example, let us discuss a policy that orders a higher quantity for each type of materials when its price is inexpensive. First, the uncertainty of the prices makes it difficult to determine the internal parameters of this policy. Although we model the prices for a calendar year as random processes, the start date and maximum duration of the project are input by users. The upper and lower limits of the prices are unknown because the start date for the project is also unknown.

Even if we determine the relevant parameters for this policy, it will probably not work well. The completed work contents for the type of activities depend not only on corresponding material, but also on the productivity of the corresponding workers and the precedence. Unsynchronized fluctuations in the weather and price curves can cause this policy to fail. For example, this policy tends to buy much concrete when the concrete price is low; however, if the weather is bad or the formwork price is high, only a small amount of concrete is consumed because of the low productivity of the workers or the undone precedence activity. Delays in progressions and waste of materials occur in this situation.

#### 6.3.2. Result and discussion

We tested the trained agents five times in scenarios #0 to #3; the project completion rate was 100% for each agent, and the average duration and cost are shown in Fig. 9. In scenario #1 (harsh budget condition), The average duration was 4.04% longer than that in scenario #0, whereas the average total cost was reduced by 2.3%, which implies that the agent learned to reduce the cost by decelerating the construction in the early stage of the project. In scenario #2, although the productivity of the workers dropped by 39.4% because of poor weather, the average duration increased only by 22.2% and the average total cost increased by 16.1%. In scenario #3 (harsh market condition), the average duration was 4.27% longer than that in scenario #0; the average total cost increased by only 1.24%.

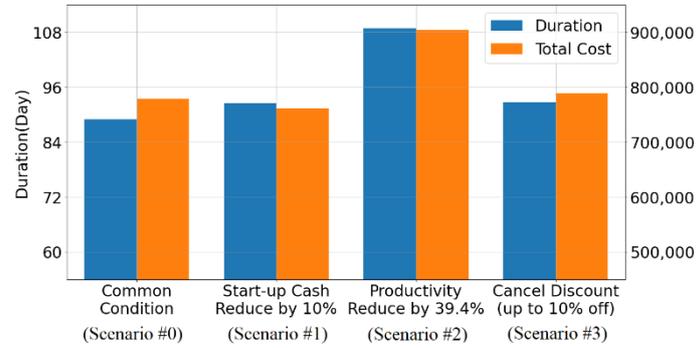

Fig. 9. Comparison of performance of the DRL-based agents under different environments

We also tested the performance of the empirical policy in scenarios #0 to #3; its comparison with the DRL agents is illustrated in Fig. 10. The project cannot be completed if we adopt the empirical policy in scenarios #1 and #2; the cost overrun occurs on the 11$^{th}$ and 30$^{th}$ days. In scenario #3, the cost overrun occurs in two out of five tests; compared with the performance of the empirical policy in scenario #0, the average total cost increased by 4.19%, whereas the average duration decreased by 0.67% in the remaining three successful tests. The performance of the empirical policy in the common condition (scenario #0) was close to the DRL-based agent, it only spent 0.67% and 3.62% more time and money, respectively. However, this experiment indicates that the advantages of DRL-based agents over empirical policy will increase if the environment becomes harsher.

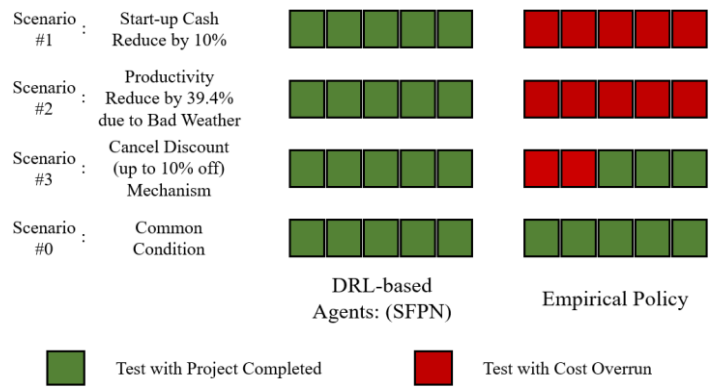

Fig. 10. Comparison of performance between the DRL-based agents and the empirical policy under different environments

### 6.4. Performance of DRL-based agents in multiple projects

#### 6.4.1. Experimental design

This experiment included one experimental and three control groups; the experimental group was conducted in scenario #0, while the control groups were conducted in scenario #4, #5 and #6, respectively. In each group, we trained and tested DRL-agents with an SFPN architecture in the corresponding scenario, and the empirical policy was the baseline. Further, we compared the advantages of the DRL-based agents over the empirical policies in different project setups.

**6.4.2. Result and discussion**

Fig. 11 shows the average performance of these agents in minimizing the total duration and cost, and the additional details are provided in Table 15. The DRL-based agents still possess remarkable capability in scenarios #4 and #5, while the advantage of the agent over empirical policy significantly increased in scenario #6. The reason for this is that the number of workers for each type changed in different proportions in scenario #6, which caused the work hours of each type of workers also changed in different proportions according to the empirical policy, further caused the unsynchronized fluctuations in the fatigue index and productivity of each type of workers, and led to the unstable workflow eventually. The results indicate that the advantage of the DRL method over empirical policy will increase if the project setups are unreasonable.

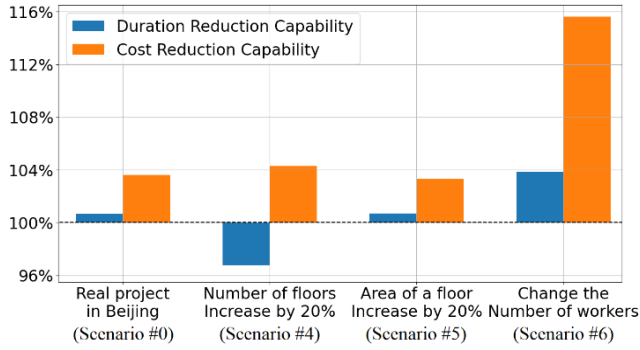

Fig. 11. Comparison between the DRL-based agents and the empirical policy in different projects

Table 15 Comparison between the DRL-based agents and the empirical policy in different projects

|  |  | Duration |  | Labor cost |  | Material cost |  | Total cost |  |
|---|---|---|---|---|---|---|---|---|---|
|  |  | Mean | Gain | Mean | Gain | Mean | Gain | Mean | Gain |
| EG: RPB | Empirical policy | 89.6 | 0% | 722.33K | 0% | 7361.13K | 0% | 8083.46K | 0% |
|  | Agent (SFPN) | 89 | −0.67% | 913.05K | 26.40% | 6877.48K | −6.57% | 7790.54K | −3.62% |
| CG #1: INF | Empirical policy | 105 | 0% | 847.66K | 0% | 8698.01K | 0% | 9545.67K | 0% |
|  | Agent (SFPN) | 108.4 | 3.24% | 1133.58K | 33.73% | 8003.34K | −7.99% | 9136.92K | −4.28% |
| CG #2: IAF | Empirical policy | 86.8 | 0% | 872.76K | 0% | 9167.34K | 0% | 10040.09K | 0% |
|  | Agent (SFPN) | 86.2 | −0.69% | 1129.52K | 29.42% | 8577.31K | -6.44% | 9706.83K | -3.32% |
| CG #3: CNWDP | Empirical policy | 98.8 | 0% | 931.47K | 0% | 8154.02K | 0% | 9085.49K | 0% |
|  | Agent (SFPN) | 95 | −3.85 | 860.87K | -7.58% | 6804.74K | -16.55% | 7665.61K | -15.63% |

\* EG and CG mean experimental and control groups, respectively; RPB, INF, IAF, and CNWDP mean real project in Beijing, increasing the number of floors, increasing the area of each floor, and changing the number of workers in different proportions, respectively

**6.5. Optimal architecture of DRL-based agent**

**6.5.1. Experimental design**

Section 3.3 described the possibility and advantage of adopting DRL-based agents with different architectures, and we have proposed agents with different architectures as shown in Fig. 7. Hence, we trained and tested all these agents in scenario #0, and compared their capabilities in duration and cost reduction.

### 6.5.1. Result and discussion

Fig. 12 shows the average performance of these agents in minimizing the total duration and cost. All five agents learned basic ideas about labor and material flows management. Even the worst agent possessed a 95% capability of the (full) empirical policy; three agents outperformed the empirical optimal policy. The best agent is the 4th agent; it is used as an SMPN to manage the material flow and follow the suggestions of the empirical work-hour policy. This reduced the total cost by 7.59% with no influence on the total duration. The additional details are provided in Table 16.

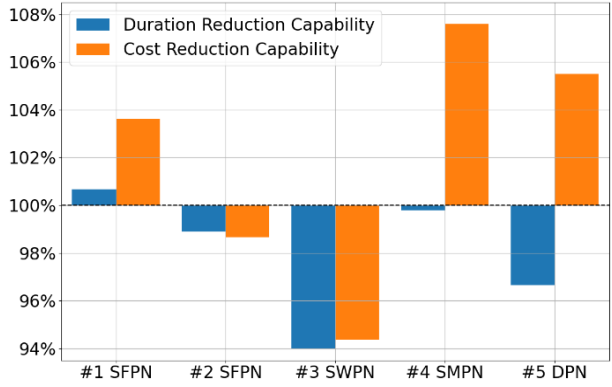

Fig. 12. Comparison between each DRL-based agent and the empirical policy

Table 16 Comparison between the DRL-based agents and empirical policy

|  | Duration | | Labor cost | | Material cost | | Total cost | |
| --- | --- | --- | --- | --- | --- | --- | --- | --- |
|  | Mean | Gain | Mean | Gain | Mean | Gain | Mean | Gain |
| Empirical policy | 89.6 | 0% | 722.33K | 0% | 7361.13K | 0% | 8083.46K | 0% |
| Agent #1: SFPN | 89 | −0.67% | 913.05K | 26.40% | 6877.48K | −6.57% | 7790.54K | −3.62% |
| Agent #2 SFPN | 90.6 | 1.12% | 808.15K | 11.88% | 7382.92K | 0.30% | 8191.06K | 1.33% |
| Agent #3: SWPN | 95 | 6.03% | 712.69K | −1.33% | 7825.04K | 6.30% | 8537.73K | 5.62% |
| Agent #4: SMPN | 89.8 | 0.22% | 723.22K | 0.12% | 6746.34K | −8.35% | 7469.60K | −7.59% |
| Agent #5: DPN | 92.6 | 3.35% | 750.28K | 3.87% | 6888.07K | −6.43% | 7638.34K | −5.51% |

We observe that the DRL algorithm outperformed the empirical policy in minimizing material cost by analyzing the records in Table 16; however, the trained SFPNs and DPN did not perform as well as the empirical policy in terms of reducing the labor cost. Although the 3rd agent spent the least budget on salary payments, its SWPN did not keep pace with the empirical material-ordering policy; this caused the duration to be five days longer and the total cost to increase. One reason for this is that the complexity and uncertainty of the material flow make it difficult for the empirical policy to cope with it, whereas the 8-hour-per-day labor allocation policy is closer to the optimal solution under the project and environmental setups of scenario #0 (scenario #0 mimics a real project, which is in common weather, budget, and market conditions). In addition, the labor cost is only a tiny part of the total cost, which makes a policy network insensitive to it. Although we can magnify the influence of labor costs by adapting the weights of the reward function, this adjustment can cause other capabilities to decrease. For example, we increased $\omega_{W,1}$ and $\omega_{W,2}$ and decreased $\omega_{M,1}$ and $\omega_{M,2}$ for the 2nd agent, which caused the labor cost to decrease; however, it increased the material cost. A DPN architecture is a better solution, and it enables the rewards of labor and material costs to not interfere with each other. Therefore, the 5th agent outperformed the first two.

### 7. Conclusion and future work

Existing optimization models in construction project management failed to address the integrated control of labor and material flows. This study thus proposes a novel model for optimizing the work and cash flows by continuous and adaptive control of the labor and material flows. The proposed model formulates the complex and uncertain interactions among the objectives and decision variables. First, coupling interactions among the work, cash, labor, material, and external flows are modelled carefully; further, multiple labor and material flows are involved in the decision variables. Finally, the uncertainty from diverse sources such as weather and market are considered for creating a robust model.

Given that our model is more complex with high uncertainty compared to the existing models, it is difficult to solve it with conventional methods. Proximal policy optimization (PPO), an online deep reinforcement learning (DRL) algorithm, is adopted to solve our model. Therefore, we first formulate our resource-flow-control model as a partially observable Markov decision process (POMDP); the POMDP includes agents that represent the resource-flow controllers and simulated scenarios that mimic real projects. The PPO algorithm trains the agents to learn the optimal resource-flow-control strategies by trial and error in the simulate scenarios.

Numerical experiments conducted in the simulated scenarios illustrates the following four facts: (1) the conventional optimization methods failed to control the labor and material flows, but our DRL-based approach succeeded; (2) the DRL-based method outperformed the empirical policy in common budget, weather, and market conditions, and had greater advantages in harsher environments; (3) the DRL-based method applied to multiple projects, and its capability remained remarkable; (4) a hybrid policy of DRL and rule-based methods, i.e., the DRL-based agent managing the material flows while the empirical policy managing the labor flows, led to the best results.

This study contributes to the intelligent management of construction projects. Our proposed POMDP-based model enhances the existing optimization models by considering complexity and uncertainty. Further, the DES can be a tool to train and test DRL-based resource-flow control agents. We designed policy and reward functions for the five agents, which can inspire other researchers. The experiments demonstrate the advantage of the DRL algorithm over other methods when optimizing complex problems in uncertain environments. Further, the trained agents can complete the projects and achieve high profit margins by avoiding wasteful and no value-adding activities, which suggests that DRL technology can empower the last planner system to achieve the goals of lean and green construction.

In the future, our resource-flow control model needs to be improved and extended. For example, current model only applies to multistory reinforced concrete buildings, more sort of projects need to be considered; control strategies for more resource flows should be studied, and this implies that we need to consider more decision variables and model interactions among more flows. More optimization objectives such as the quality of the project and the satisfaction of the stakeholders should be considered. Further, larger/deeper and more effective DNN architecture will be proposed and tested, because the more complex models need to be solved. Finally, we will try to employ more DRL algorithms for improving the training effects and reducing the training time.

## Acknowledgments

This study was supported by a grant from Glodon Company Limited and the Research Development Project of the Ministry of Housing and Urban-Rural Development of the People's Republic of China (K20210032).

## Appendix A. External flows module of the transition function

The external-flows module first generates baseline curves based on historical statistics. For the weather information, the baseline curves consider piecewise fittings of the average temperature, probability of precipitation, monthly precipitation, and wind speed curves over the years in Beijing (raw curves are shown on the website [67]). For the material prices, the baseline curves equal the product of trigonometric and linear functions; the trigonometric functions represent periodicity, whereas the linear functions represent inflation. The base price, peak dates, and inflation rates are determined based on the news records in China. The base prices of rebar, formwork, and concrete are 3,000 CNY/ton, 20 CNY/m$^2$, and 480 CNY/m$^3$, respectively; the peak dates of the rebar and formwork prices are approximately mid-September and mid-April, respectively, whereas the periodic fluctuations of concrete price are not obvious; the inflation rates for rebar, formwork, and concrete are 30%, 5%, and 10%, respectively.

Then, the external-flows module samples annual curves of weather ($Tp_t$, $Rf_t$, and $Ws_t$) and price ($RbPr_t$, $FwPr_t$ and $CcPr_t$) states at the beginning of each simulation. For the weather information, the rainy dates are determined based on the probability of precipitation baseline, and then, the annual curve of $Rf_t$ is determined based on the average monthly precipitation baseline and rainy dates; the annual curves of $Tp_t$ and $Ws_t$ are finally determined by adjusting the baseline of the temperature and wind speed, while assuming that the temperature rises and wind speed drops before the rainy day. A reverse situation occurs on a rainy day, and the increments and decrements are positively correlated with $Rf_t$. The examples of the sampled annual curves for $Rf_t$, $Tp_t$, and $Ws_t$ are indicated in Fig. A.1. For the material prices, the annual curves of $RbPr_t$, $FwPr_t$, and $CcPr_t$ are determined by adding minor white noise processes to their baselines.

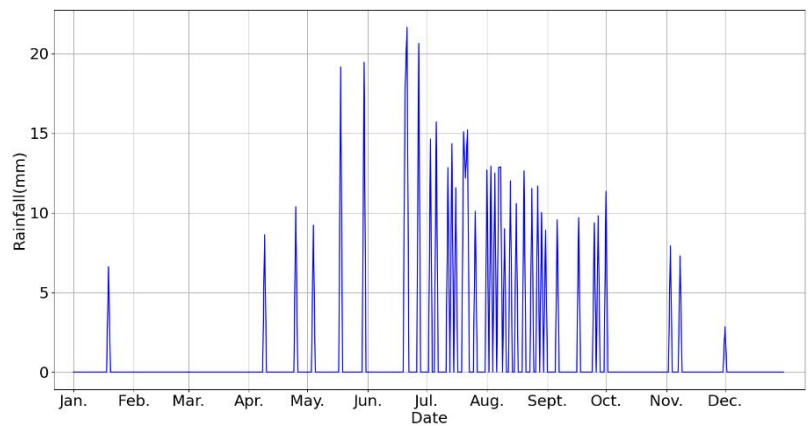

(a)

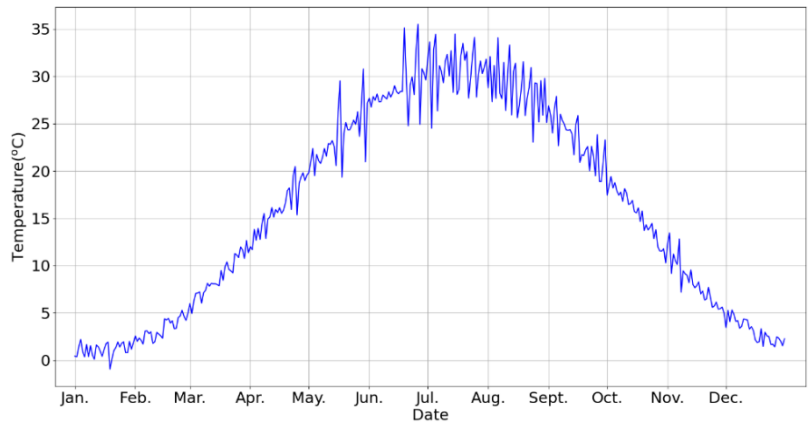

(b)

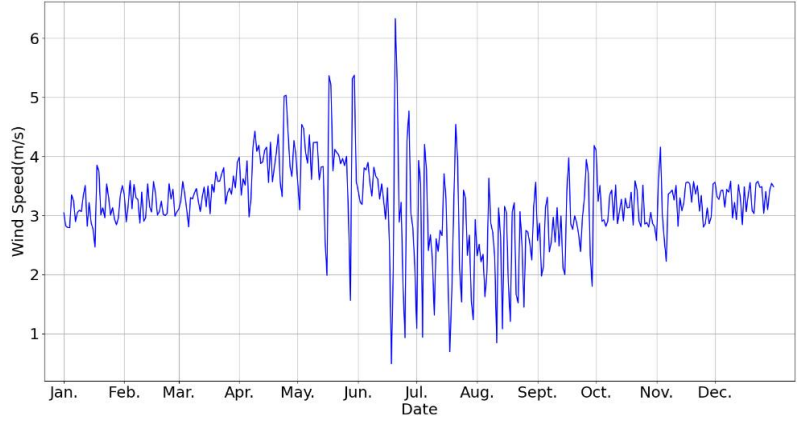

(c)

Fig. A.1. Examples of generated annual curves of (a) $Rf_t$, (b) $Tp_t$, (c) $Ws_t$

## Appendix B. Forecast module of the observation function

The forecast module mimics the forecast of cash inflow, weather, and material prices in the real world. For the cash inflow, the module predicts $ICa_{t+1}$ to $ICa_{t+3}$ based on the progressions because the milestone payments are the main part of cash inflow. For the weather forecast, the module first predicts the rainfall within the next three days; the nearer the day, the higher is the accuracy. Further, the module predicts the temperature and wind speed by adding noise to the value from $Tp_{t+1}/Ws_{t+1}$ to $Tp_{t+3}/Ws_{t+3}$, and the noises are positively correlated with the rainfall prediction error. For the material price forecast, the module adds noise to the value from $Pr_{t+1}$ to $Pr_{t+5}$; the closer the day, the smaller is the noise.

## Appendix C. Empirical optimal policy

We select a static policy adopted by human managers in the real project in Beijing; we call it the empirical optimal policy. The main principle of the empirical policy is to advance the progressions steadily; i.e., the policy requests each type of worker to advance the progression of the corresponding construction activity at a speed of 1/3 floors per day. This principle theoretically avoids the precedence flow from becoming the critical flow (workers are delayed by the incomplete status of their previous activity in this case).

According to the main principle, the number of work hours of each type of workers can be calculated with

$$WH_t = \left\lfloor \frac{\text{pFZ} \cdot \text{pZA}}{3 \cdot \text{pWH2A}} \right\rfloor + 1 . \tag{C.1}$$

For the materials, the quantities of rebars and concrete ordered per day are equal to the amount consumed, which can be calculated with

$$B_t = \left\lfloor \frac{\text{pFZ} \cdot \text{pZA}}{3 \cdot \text{pS2A}} \right\rfloor + 1 . \tag{C.2}$$

The formworks can be recycled, and therefore, the manager does not need to buy them on a daily basis. However, the manager must ensure that the stock of the formworks is higher than the value calculated with Eq. (C.2). If the stock falls below the value because of wear and tear, the manager will need to replenish 50% at one time.